\documentclass[a4paper, 11pt]{article}
\pdfoutput=1  
\usepackage{etoolbox,verbatim}
\usepackage{wrapfig}
\newtoggle{arxiv}
\toggletrue{arxiv} 

\newtoggle{customthms}
\toggletrue{customthms} 

\newtoggle{colt}
\togglefalse{colt} 

\usepackage{xcolor}
\usepackage{tcolorbox}
\tcbuselibrary{skins,breakable}

\tcbset{
  aibox/.style={
      width=\linewidth,
      top=6pt,
      bottom=0pt,
      left=1.5pt,
      right=1.5pt,
      colback=blue!60!gray!5,
      colframe=black,
      colbacktitle=black,
      enhanced,
      center,
      attach boxed title to top left={yshift=-0.1in,xshift=0.15in},
      boxed title style={boxrule=0pt,colframe=white,},
    }
}
\newtcolorbox{AIbox}[2][]{aibox,title=#2,#1}

\definecolor{primalcolor}{HTML}{A60000}
\definecolor{contrarycolor}{HTML}{00A6A6}
\definecolor{darkcontrarycolor}{HTML}{004C4C}
\definecolor{lightblue}{HTML}{2970CC}
\definecolor{lightpurple}{HTML}{673147}
\definecolor{ForestGreen}{HTML}{FF5733}
\definecolor{myred}{HTML}{AA4A44}
\definecolor{hyppurple}{HTML}{800080}

\newcommand{\linkcolor}{darkcontrarycolor}
\newcommand{\urlcolor}{darkcontrarycolor}
\newcommand{\citecolor}{darkcontrarycolor}

\newcommand{\thmcolor}{red!40!black}
\newcommand{\thmcolordark}{red!30!black}

\usepackage[T1]{fontenc}
\usepackage{courier}
\usepackage{capt-of}
\usepackage[font=small,labelfont=bf]{caption}
\usepackage{scalefnt}

\usepackage{natbib}
\usepackage{scalefnt}
\usepackage{amssymb,upgreek,bm}
\usepackage{mathtools}
\usepackage[english]{babel}  
\usepackage{multirow,tcolorbox,tikz-cd,turnstile,xspace,xparse}
\usepackage{relsize,breakcites,appendix}
\usepackage{longtable,tablefootnote,booktabs,float,makecell}
\usepackage[normalem]{ulem}
\usepackage{tabu}
\usepackage[shortlabels]{enumitem} 
\usepackage{footnote}
\usepackage{boxedminipage}
\usepackage{multirow,nicefrac}
\usepackage{verbatim} 

\usepackage[colorlinks=true,linkcolor=\linkcolor,urlcolor=\urlcolor,citecolor=\citecolor,breaklinks]{hyperref}

\usepackage{prettyref}

\usepackage[capitalise,nameinlink]{cleveref}

\iftoggle{colt}{
    \DeclareRobustCommand{\qed}{
        \usepackage{thmtools}
          \ifmmode \mathqed
          \else
            \leavevmode\unskip\penalty9999 \hbox{}\nobreak\hfill
            \quad\hbox{\qedsymbol}%
          \fi
    }
}
{
    \usepackage{amsthm}
     
}
\usepackage{thm-restate}

\iftoggle{arxiv}
{
\usepackage{mathrsfs}

    \usepackage[
  letterpaper,
  left=1in,
  right=1in,
  top=0.9in,
  bottom=0.9in,
  headheight=14pt,
  headsep=14pt,
  footskip=20pt
]{geometry}

\usepackage{fancyhdr}
\usepackage{lipsum} 

\pagestyle{fancy}
\fancyhf{} 
\fancyfoot[C]{\small \thepage}

\renewcommand{\headrulewidth}{0.4pt}
\renewcommand{\footrulewidth}{0pt}

    \usepackage[bitstream-charter]{mathdesign}
     
    \usepackage[scaled=0.92]{PTSans}

    \usepackage{subfig}

    \usepackage[noend]{algpseudocode}
    \usepackage{algorithm}
   
}
{
    \usepackage{mathrsfs}
}

\DeclareMathAlphabet{\mathbfsf}{\encodingdefault}{\sfdefault}{bx}{n}

\numberwithin{equation}{section}

\iftoggle{arxiv}
{
    \newcommand{\colorbold}[1]{
    \textbf{\textcolor{\thmcolor}{#1}}}}
{
    \newcommand{\colorbold}[1]{\textbf{#1}
}
}
\iftoggle{arxiv}
{
    
}
{
    
}

\usepackage{thmtools}

\Crefname{equation}{Eq.}{Eqs.}
\Crefname{assumption}{Assumption}{Assumptions}
\Crefname{condition}{Condition}{Conditions}
\Crefname{claim}{Claim}{Claims}
\Crefname{property}{Property}{Properties}
\Crefname{construction}{Construction}{Constructions}

\declaretheoremstyle[
    headformat=\normalfont\textcolor{\thmcolordark}{\bfseries\NAME\,\NUMBER}\NOTE,%
    notefont={\normalfont\textcolor{\thmcolordark}{\bfseries}}, 
    notebraces={}{},
    bodyfont=\normalfont\itshape,
    spaceabove = 6pt,
    spacebelow = 6pt,
    ]{coloredthmversion}

\declaretheoremstyle[
    headformat=\normalfont\textcolor{\thmcolordark}{\bfseries\NAME\,\NUMBER}\NOTE,%
    bodyfont=\normalfont\itshape,
    spaceabove = 6pt,
    spacebelow = 6pt,
    ]{coloredthm}

\declaretheoremstyle[
    headformat=\normalfont\textcolor{\thmcolordark}{\bfseries\NAME\,\NUMBER}\NOTE,%
    bodyfont=\normalfont,
    spaceabove = 6pt,
    spacebelow = 6pt,
    ]{coloreddef}

\iftoggle{customthms}
{
    \theoremstyle{coloredthmversion}
}
{}

\iftoggle{customthms}{
  \theoremstyle{coloredthm}
  \newtheorem{theorem}{Theorem}
  \newtheorem{lemma}{Lemma}[section]
  \newtheorem{corollary}{Corollary}[section]
  \newtheorem{proposition}[lemma]{Proposition}
}
{}

\newtheorem*{thminformal*}{Informal Theorem}

\iftoggle{customthms}{
    \theoremstyle{coloreddef}
    \newtheorem{definition}{Definition}[section]

    \newtheorem{property}{Property}[section]
}
{
  
}

\newtheorem{assumption}{Assumption}[section]
\newtheorem{condition}{Condition}[section]

\makeatletter
\newcommand{\neutralize}[1]{\expandafter\let\csname c@#1\endcsname\count@}
\makeatother

\iftoggle{customthms}{
    \newtheoremstyle{named}{}{}{\itshape}{}{\bfseries}{}{.5em}{\Cref{#3} {\normalfont (informal)} }{}
    \theoremstyle{named}
    
    \theoremstyle{plain}
}
{}

\newtheorem*{theorem*}{Theorem}
\newtheorem*{lemma*}{Lemma}
\newtheorem*{corollary*}{Corollary}
\newtheorem*{proposition*}{Proposition}
\newtheorem*{claim*}{Claim}
\newtheorem*{fact*}{Fact}
\newtheorem*{observation*}{Observation}
\newtheorem*{definition*}{Definition}
\newtheorem*{remark*}{Remark}
\newtheorem*{example*}{Example}

\def\bv{\mathbf{v}}

\def\ddefloop#1{\ifx\ddefloop#1\else\ddef{#1}\expandafter\ddefloop\fi}
\def\ddef#1{\expandafter\def\csname bb#1\endcsname{\ensuremath{\mathbb{#1}}}}
\ddefloop ABCDEFGHIJKLMNOPQRSTUVWXYZ\ddefloop

\def\ddefloop#1{\ifx\ddefloop#1\else\ddef{#1}\expandafter\ddefloop\fi}
\def\ddef#1{\expandafter\def\csname frak#1\endcsname{\ensuremath{\mathfrak{#1}}}}
\ddefloop ABCDEFGHIJKLMNOPQRSTUVWXYZ\ddefloop

\def\ddefloop#1{\ifx\ddefloop#1\else\ddef{#1}\expandafter\ddefloop\fi}
\def\ddef#1{\expandafter\def\csname fr#1\endcsname{\ensuremath{\mathfrak{#1}}}}
\ddefloop ABCDEFGHIJKLMNOPQRSTUVWXYZ\ddefloop

\def\ddefloop#1{\ifx\ddefloop#1\else\ddef{#1}\expandafter\ddefloop\fi}
\def\ddef#1{\expandafter\def\csname eul#1\endcsname{\ensuremath{\EuScript{#1}}}}
\ddefloop ABCDEFGHIJKLMNOPQRSTUVWXYZ\ddefloop

\def\ddefloop#1{\ifx\ddefloop#1\else\ddef{#1}\expandafter\ddefloop\fi}
\def\ddef#1{\expandafter\def\csname scr#1\endcsname{\ensuremath{\mathscr{#1}}}}
\ddefloop ABCDEFGHIJKLMNOPQRSTUVWXYZ\ddefloop

\def\ddefloop#1{\ifx\ddefloop#1\else\ddef{#1}\expandafter\ddefloop\fi}
\def\ddef#1{\expandafter\def\csname b#1\endcsname{\ensuremath{\mathbf{#1}}}}
\ddefloop ABCDEFGHIJKLMNOPQRSTUVWXYZ\ddefloop

\def\ddefloop#1{\ifx\ddefloop#1\else\ddef{#1}\expandafter\ddefloop\fi}
\def\ddef#1{\expandafter\def\csname bhat#1\endcsname{\ensuremath{\hat{\mathbf{#1}}}}}
\ddefloop ABCDEFGHIJKLMNOPQRSTUVWXYZ\ddefloop

\def\ddefloop#1{\ifx\ddefloop#1\else\ddef{#1}\expandafter\ddefloop\fi}
\def\ddef#1{\expandafter\def\csname btil#1\endcsname{\ensuremath{\tilde{\mathbf{#1}}}}}
\ddefloop ABCDEFGHIJKLMNOPQRSTUVWXYZ\ddefloop

\def\ddefloop#1{\ifx\ddefloop#1\else\ddef{#1}\expandafter\ddefloop\fi}
\def\ddef#1{\expandafter\def\csname bst#1\endcsname{\ensuremath{\mathbf{#1}^\star}}}
\ddefloop ABCDEFGHIJKLMNOPQRSTUVWXYZ\ddefloop

\def\ddefloop#1{\ifx\ddefloop#1\else\ddef{#1}\expandafter\ddefloop\fi}
\def\ddef#1{\expandafter\def\csname bst#1\endcsname{\ensuremath{\mathbf{#1}^\star}}}
\ddefloop abcdeghijklmnopqrstuvwxyz\ddefloop

\def\ddefloop#1{\ifx\ddefloop#1\else\ddef{#1}\expandafter\ddefloop\fi}
\def\ddef#1{\expandafter\def\csname bhat#1\endcsname{\ensuremath{\hat{\mathbf{#1}}}}}
\ddefloop abcdefghijklmnopqrstuvwxyz\ddefloop

\def\ddefloop#1{\ifx\ddefloop#1\else\ddef{#1}\expandafter\ddefloop\fi}
\def\ddef#1{\expandafter\def\csname b#1\endcsname{\ensuremath{\mathbf{#1}}}}
\ddefloop abcdeghijklnopqrstuvwxyz\ddefloop

\def\ddefloop#1{\ifx\ddefloop#1\else\ddef{#1}\expandafter\ddefloop\fi}
\def\ddef#1{\expandafter\def\csname barb#1\endcsname{\ensuremath{\bar{\mathbf{#1}}}}}
\ddefloop abcdefghijklmnopqrstuvwxyz\ddefloop

\def\ddef#1{\expandafter\def\csname c#1\endcsname{\ensuremath{\mathcal{#1}}}}
\ddefloop ABCDEFGHIJKLMNOPQRSTUVWXYZ\ddefloop
\def\ddef#1{\expandafter\def\csname h#1\endcsname{\ensuremath{\widehat{#1}}}}
\ddefloop ABCDEFGHIJKLMNOPQRSTUVWXYZ\ddefloop
\def\ddef#1{\expandafter\def\csname hc#1\endcsname{\ensuremath{\widehat{\mathcal{#1}}}}}
\ddefloop ABCDEFGHIJKLMNOPQRSTUVWXYZ\ddefloop
\def\ddef#1{\expandafter\def\csname t#1\endcsname{\ensuremath{\widetilde{#1}}}}
\ddefloop ABCDEFGHIJKLMNOPQRSTUVWXYZ\ddefloop
\def\ddef#1{\expandafter\def\csname tc#1\endcsname{\ensuremath{\widetilde{\mathcal{#1}}}}}
\ddefloop ABCDEFGHIJKLMNOPQRSTUVWXYZ\ddefloop

\usepackage{tcolorbox}
\tcbuselibrary{skins,breakable}

\tcbset{
  aibox/.style={
      width=\linewidth,
      top=6pt,
      bottom=0pt,
      left=1.5pt,
      right=1.5pt,
colback=blue!60!gray!5,
colframe=black,
      colbacktitle=black,
      enhanced,
      center,
      attach boxed title to top left={yshift=-0.1in,xshift=0.15in},
      boxed title style={boxrule=0pt,colframe=white,},
    }
}

\iftoggle{arxiv}
{

}
{

}

\iftoggle{arxiv}
{

}
{
\newcommand{\textbf}[1]{\colopar{#1}}
}

\newcommand{\ballkr}[1][r]{\cB_{k}(r)}

\DeclareMathSymbol{\shortminus}{\mathbin}{AMSa}{"39}

\newcommand{\R}{\mathbb{R}}

\usepackage{preamble/color-edits}

\addauthor{ms}{magenta}

\newcommand{\ignore}[1]{}

\iftoggle{arxiv}
{
\input{preamble/arxiv_title}
}
{}

\title{\textcolor{\thmcolor}{Nano World Models}: A Minimalist Implementation of Future Video Prediction}

\author{
  \small Siqiao Huang\footnote{\texttt{huang-sq23@mails.tsinghua.edu.cn} \label{foot:tsing}}\textsuperscript{,$\dagger$}  ~ Partha Kaushik\footnote{\texttt{\{parthak,mchen5,hengkaip,kgeng, ochehab,msimchow\}@andrew.cmu.edu}\label{foot:cmu}} ~ Michael Chen\textsuperscript{\ref{foot:cmu}} ~ Hengkai Pan\textsuperscript{\ref{foot:cmu}}  ~ Kaiwen Geng\textsuperscript{\ref{foot:cmu}}\\
  \small Omar Chehab\textsuperscript{\ref{foot:cmu}} ~ Fernando Moreno-Pino\textsuperscript{$c,d$} ~ Max Simchowitz\textsuperscript{\ref{foot:cmu},e} ~ \\
  \vspace{-.3em}
 \rule{.38\textwidth}{.7pt}\\
   \footnotesize $^a$Tsinghua University~~$^b$Carnegie Mellon University \\
   \footnotesize $^c$University of Bristol~~ $^d$University of Oxford~~~$^e$Amazon FAR \\
   \footnotesize $^\dagger$Project Lead\\
}
\papercode{https://github.com/simchowitzlabpublic/nano-world-model}
\paperblog{https://simchowitzlabpublic.github.io/nano-world-model}
\papermodels{https://huggingface.co/collections/knightnemo/nano-world-model}
\paperdate{\today}

\date{\vspace{-.5em}}
\begin{document}

\begin{tcolorbox}[
colback=blue!60!gray!5, colframe=gray!50, 
boxrule=0pt, 
    arc=2mm
]
\maketitle
\tcbline
\vspace{-.7cm}
\newcommand{\NWM}{\colorbold{Nano World Models}}

\begin{minipage}[t]{1.0\linewidth}
    \centering
\includegraphics[width=\textwidth]{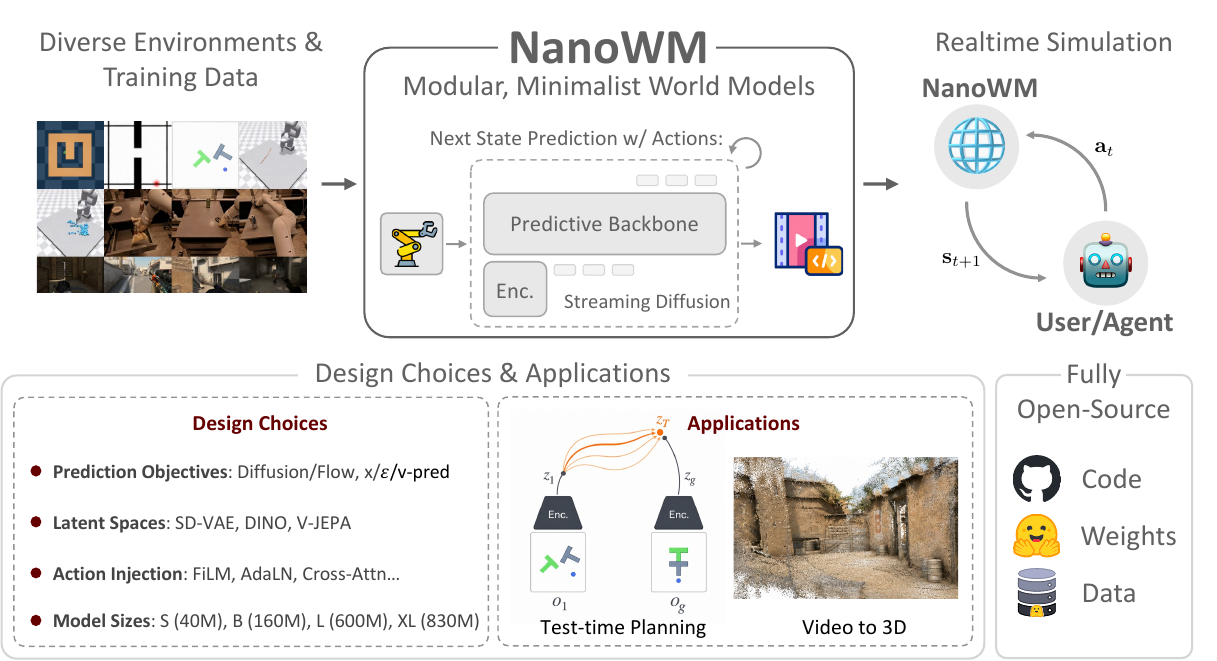}

             \captionof{figure}{
        \small  \textbf{Overview.} \NWM{} is a minimalist and modular framework for future video prediction and world modeling. It supports diverse environments and training data, encodes observations into latent spaces, and predicts future observations with a shared diffusion-forcing interface that can accommodate different objectives, model sizes, and action-conditioning mechanisms. The same model interface enables realtime simulation, test-time planning, and video-to-3D applications, while the project fully open-sources code, model weights, and data to support reproducible study of world-model design choices.} 
    \label{fig:teaser}
\end{minipage}
  \vskip 0.5cm
  \makeatletter
  \tcbline 
  \vspace{-.7cm}
  \ifdefempty{\metadatalist}{}{\metadatalist\par}
  \makeatother
  
\end{tcolorbox}
\begin{abstract}
\newcommand{\NWM}{\colorbold{Nano World Models}}

World models have become a central paradigm for learning predictive simulators that support generation, planning, and decision-making. Yet, despite rapid progress in industry-scale interactive video generation, the broader research community still lacks compact, reproducible, and easily extensible implementations for studying the design choices underlying modern world models. We introduce \NWM{}, a minimalist codebase for future video prediction centered around diffusion forcing. \NWM{} provides a unified interface for generative objectives, model scales, action-conditioning mechanisms, latent observation spaces, datasets, evaluation protocols, and long-horizon rollout procedures. This design enables controlled studies of world-modeling components that are often entangled across separate implementations. Through experiments across simple control environments, game simulation, and real-robot data, we examine how prediction parameterization, architecture scale, action injection, sampling budget, and domain complexity affect video prediction quality and autoregressive rollout behavior. By releasing code, configurations, evaluation scripts, and pretrained checkpoints, \NWM{} aims to provide a compact yet extensible experimental substrate for open, reproducible, and scientific world-model research.

\end{abstract}

\newcommand{\beps}{\bm{\epsilon }}
\newcommand{\NWM}{\colorbold{Nano World Models}}
\section{Introduction}
World Models \citep{ha2018recurrent,dawid2023introduction} have emerged as a cornerstone of spatial intelligence \citep{yang2025cambrian, wang2026vagen} and real-world decision-making \citep{richens2025general, guo2025ctrl}, generating high-fidelity futures by conditioning on the agent's history and actions.  Especially in the past few months predating the release of this manuscript, we have witnessed significant advances in industry-scale World Models \citep{parkerholder2025genie3,lingbot-world}. Yet, for the broader community, the gap between reading about these models and deploying them remains disappointingly wide.

This manuscript accompanies \NWM: a minimalist, batteries-included repository for advancing a careful and scientific approach to world-model design. The motivation for this project is simple: while industry-scale world models achieve stunning visual effects, they are built around a handful of simple, well-established techniques: video diffusion \citep{ho2022video, svd}, diffusion forcing \citep{chen2024diffusion}, consistency distillation \citep{song2023consistency} etc.

We posit that as world model algorithm stabilizes, a shift in research focus lies from inventing new techniques to developing a more nuanced understanding of subtler scientific design decisions, including architectural choices, training objectives, and, of course, data composition and scaling behavior. However, one cannot truly understand the science behind a model without being able to easily experiment with it. Especially with the current fragmented landscape of world modeling research, diverse datasets \citep{brohan2022rt,pearce2022counter,zhou2024dino}, training recipes \citep{yang2023diffusion,lipman2024flow, li2025back}, evaluation protocols \citep{vafa2024world,zhang2026worldinworld} and downstream tasks \citep{alonso2024diffusion, quevedo2026worldgym, guo2025ctrl} scattered across numerous sources makes rigorous scientific studies extremely hard.

\subsection{Contributions.}

We introduce \NWM{}, a minimalist, batteries-included implementation of world models, as a usable starting point admist the aforementioned fragmented landscape. Our repo is built around specializing the diffusion-forcing \citep{chen2024diffusion} model.  We enable hydra-based configurations \citep{Yadan2019Hydra}, modular code for data loading, model design, training, evaluation, and downstream tasks, as well as well-curated docs. In greater detail, we support the following features:
\begin{itemize}
    \item \textbf{Generative Modeling Objectives.} Centered around Diffusion Forcing \citep{chen2024diffusion}, \NWM{} supports a variety of generative modeling and prediction objectives. It supports Diffusion \citep{rombach2022high,peebles2023scalable} and Flow-Matching \citep{lipman2022flow,lipman2024flow}, with $\bx$, $\beps$ and $\mathbf{v}$ prediction objectives \citep{li2025back}.
    \item \textbf{Architectural Sizes and Design.} \NWM{} is built for support of architectures of varying sizes. Following naming conventions from the image generation community \citep{ma2024sit,peebles2023scalable}, we include four variants of varying sizes: NanoWM-S (40M), NanoWM-B (160M), NanoWM-L (600M) and NanoWM-XL (830M). For action injection methods, \NWM{} supports element-wise addition, AdaLN \citep{huang2017arbitrary}, AdaLN fused with timestep injection, FiLM \citep{perez2018film} and cross-attention.
    \item \textbf{Environments.} \NWM{} supports diverse environments, ranging from simple simulation environments to game simulation and robot manipulation. For simple simulation environments, \NWM{} currently supports environments drawn from standard robotics benchmarks, namely D4RL \citep{fu2020d4rl} and DeepMind Control Suite \citep{tassa2018deepmind}. These environments include maze navigation (\texttt{Maze}, \texttt{Wall}), fine-grained control for tabletop pushing (\texttt{PushT}) and deformable object manipulation with an XArm (\texttt{Rope}, \texttt{Granular}). For game simulation, we support the well-celebrated CS:GO dataset \citep{pearce2022counter} and for robot manipulation, we support the widely used RT-1 \citep{brohan2022rt} dataset.
    \item \textbf{Logging and Evaluations.} \NWM{} supports both Tensorboard \citep{abadi2016tensorflow} and Wandb \citep{wandb} logging systems. Loggings include callback-style validation, reliable per-step checkpointing and system utilization informations. Evaluations are fixed-seed reproducible.
    \item \textbf{Long-Horizon Generation.} \NWM{} goes beyond the training context. Empowering by the auto-regressive capability of diffusion forcing, as well as sliding window approaches, \NWM{} can produce temporally consistent long video generations with 4x the training horizon.
\end{itemize}

\subsection{World-Modeling as\colorbold{Tool-Use}}

\NWM{} goes beyond next state prediction, supporting multiple applications where world modeling serves as a tool for downstream tasks.
\begin{itemize}
    \item \textbf{3D Scene Generation.} Beyond serving as a video predictor, a world model can also act as a generative prior for constructing 3D-consistent scenes. \NWM{} supports exporting generated video rollouts into downstream 3D pipelines \citep{depthanything3,chen2026geometric}, where multi-view or temporally adjacent predictions can be lifted into scene representations such as point clouds, Gaussian splats, or neural fields. This provides a lightweight bridge between 2D video generation and 3D scene synthesis, extracting generations from video world models to persistent 3D scenes.
\item \textbf{Goal-Conditioned Planning.}
\NWM{} further supports goal-conditioned planning, where the world model is used as a simulator for evaluating candidate action sequences before execution. Given an initial observation and a desired goal state, the model can roll out possible futures under different action proposals, enabling planning by trajectory optimization. This turns the learned dynamics model into a tool for decision-making: rather than directly learning a policy for every task, users can query the model to imagine, compare, and select action sequences that are most likely to reach the goal.

\end{itemize}

\subsection{Mission Statement}

Thanks to its modular design, \NWM{} makes experimentation a matter of changing modular configs rather than rewriting pipelines. Datasets, tasks, model sizes, prediction objectives and overriding any specific configuration, can be swapped from a single command line. In addition, we release everything: code, data and more than a dozen pretrained model checkpoints of all sizes.
World Models need the World. Our hope is to build a Babel tower for world model research: datasets, objectives, architectures, and tasks all speaking the same language. We invite the global community to join us, contribute, and build this future together. Learn more:
\begin{itemize}
    \item[{\includegraphics[height=1em]{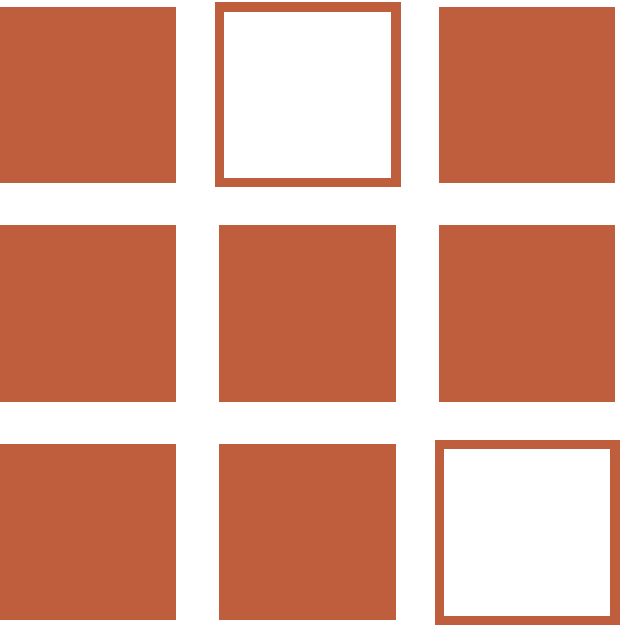}}] \textbf{Blog:} {\small\url{https://simchowitzlabpublic.github.io/nano-world-model}}
    \item[{\includegraphics[height=1em]{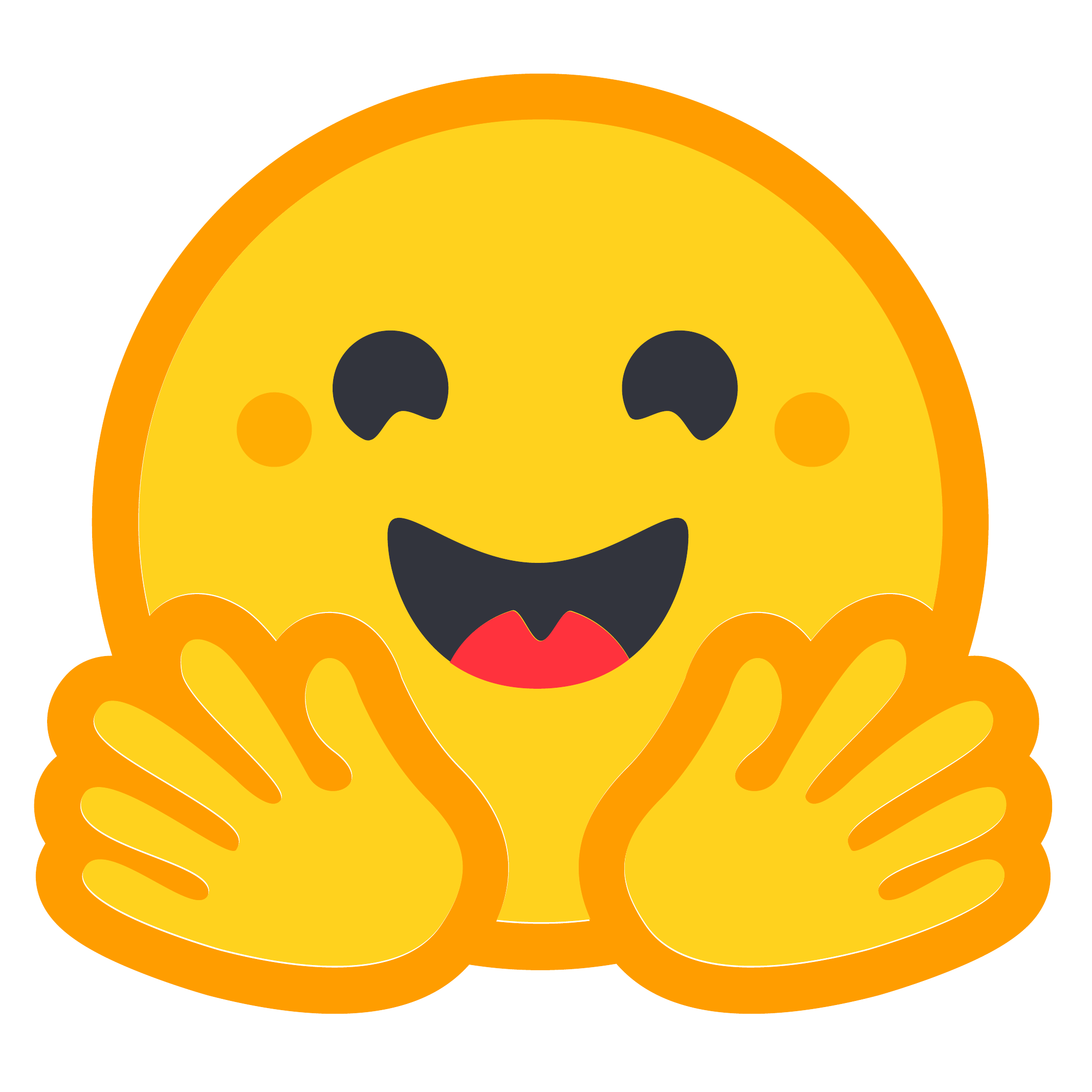}}] \textbf{Models:} {\small\url{https://huggingface.co/collections/knightnemo/nano-world-model}}
    \item[{\includegraphics[height=1em]{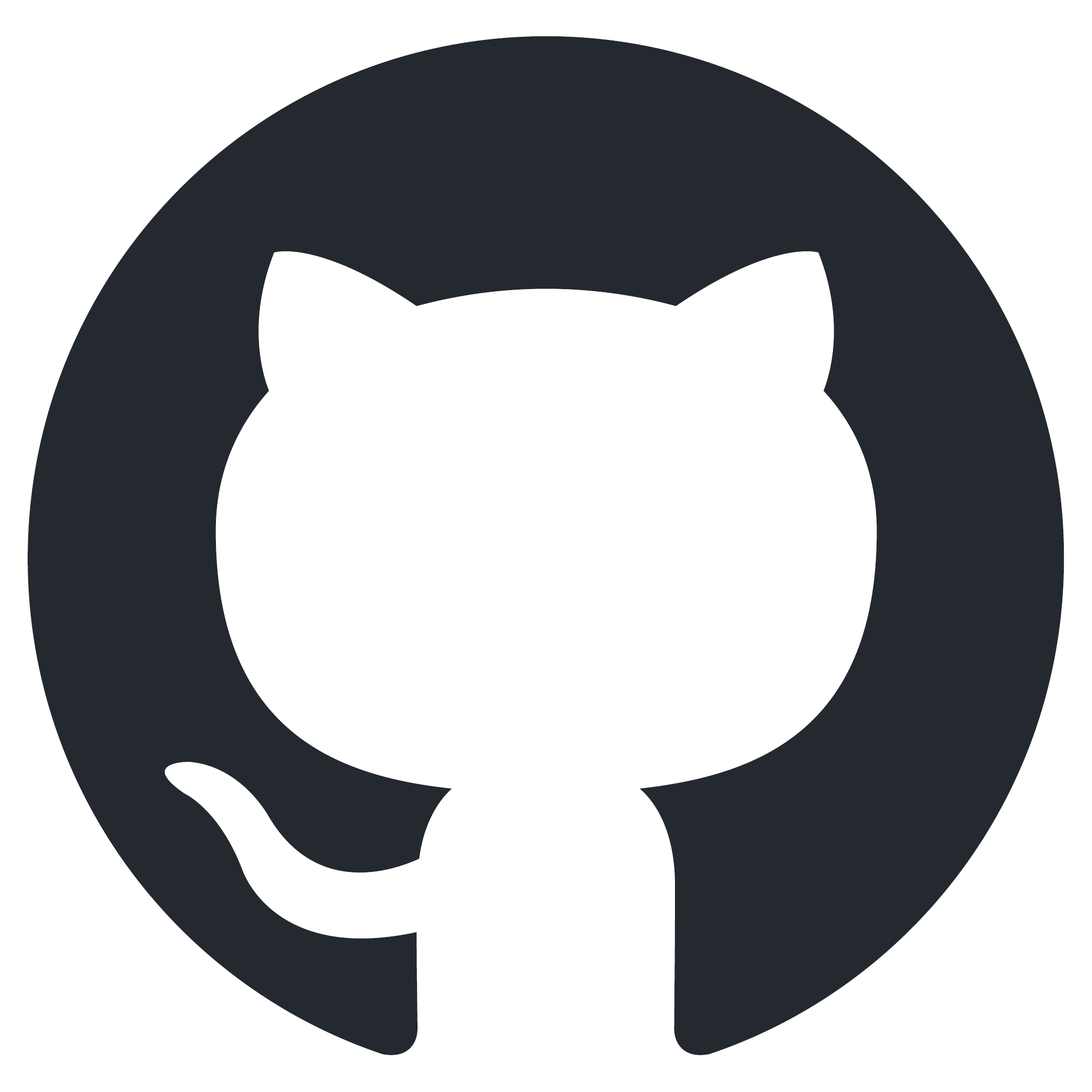}}] \textbf{Code:} {\small\url{https://github.com/simchowitzlabpublic/nano-world-model}}
\end{itemize}


\newcommand{\enc}{\mathrm{enc}}

\section{Preliminaries}
\label{sec:prelims}
We define \textbf{world-modeling} as the problem of modeling posterior distributions over sequences of high dimensional observations. Specifically, we are given a sequence $\bo_1,\dots,\bo_{T}$ of previous observations (e.g. future video frames), as well as a conditioning variable $c$ (e.g. a text-description) and our goal is to produce a completion $\bo_{T+1},\dots,\bo_{T+H}$ of future observations. In many cases, we do not predict observations directly, but instead predict \textbf{encodings} $\bx_t := \enc_{\psi}(\bo_t)$, where $\enc_{\psi}$ is a learned or pretrained encoder (e.g. VAE \citep{kingma2013auto}, DINO \citep{caron2021emerging} or V-JEPA \citep{bardes2024revisiting}). Our aim is to predict the conditional distribution of sequences of the conditional variable. We view this task as a generative model problem, where we assume there is some true distribution $p^\star(\bx_{T+1:T+H} \mid \bx_{1:T},c)$, and we learn a probabilistic model $p_{\theta}(\cdot \mid \cdot)$ over these conditionals. We denote samples from this model as $\bhatx_{T+1:T+H} \sim p_{\theta}(\cdot \mid \bx_{1:T},c)$.

\paragraph{Planning with World Models}
Given a sequence of candidate actions $\ba_{T:T+H-1}$, and the world model can be used as a simulator for evaluating their likely consequences. Given a task-specific utility or reward function $R(\bx_{T+1:T+H}, \ba_{T:T+H-1})$, planning amounts to searching for an action sequence whose predicted rollout has high expected return under the learned model:
\[
    \ba^\star_{T:T+H-1}
    \in
    \arg\max_{\ba_{T:T+H-1} \in \mathcal{A}^H}
    \mathbb{E}_{\bhatx_{T+1:T+H} \sim
    p_{\theta}(\cdot \mid \bx_{1:T}, c, \ba_{T:T+H-1})}
    \left[
        R(\bhatx_{T+1:T+H}, \ba_{T:T+H-1})
    \right].
\]
In practice, this optimization is typically performed approximately, for example by sampling or optimizing a population of candidate action sequences using model predictive control (MPC). At each decision step, the planner rolls out the world model over a finite horizon, selects the best candidate sequence according to the predicted return, executes only the first action, observes the next state, and replans.








\section{Methods Supported: Unification through Diffusion Forcing}
\label{sec:methods}

Having defined world modeling as conditional sequence generation in
Section~\ref{sec:prelims}, we now describe the modeling and software
abstractions powering \NWM{}. The central design principle is to
treat diffusion forcing as a unified interface: prediction objectives,
architectures, action-conditioning mechanisms, latent spaces, environments, and
rollout procedures can be exchanged while preserving the same training and
sampling pipeline.

\subsection{Diffusion Forcing as a Unified Interface}

Popular generative models produce samples through iterative computation.
Diffusion models iteratively denoise corrupted samples, while flow-matching
models learn a vector field that transports a simple base distribution to the
data distribution. Diffusion forcing extends this view to sequence modeling by
allowing different frames in the same trajectory to occupy different stages of
the generation process.

We introduce a noise index set $\bbK \subset \R$. For an encoded trajectory
$\bx_{1:T+H}$, diffusion forcing assigns each frame $\bx_t$ a noise index
$k_t \in \bbK$. The model is trained on noised trajectories together with their
noise-index schedule $\bk = (k_1,\dots,k_{T+H})$. Context frames may be kept clean
or nearly clean, while future frames may be assigned higher-noise indices. By
changing only this schedule, Nano World Models can express teacher-forced
prediction, masked future prediction, and autoregressive rollout using the same
model interface.

\subsection{Generative Objectives}

Nano World Models supports multiple generative objectives under the same
diffusion-forcing interface. For diffusion objectives, the model can be trained
with $\bx$-prediction, $\beps$-prediction, or $\bv$-prediction targets \citep{li2025back}. For
flow-matching objectives, the model predicts the velocity field induced by a
chosen interpolant, such as the linear interpolant between data and noise.

Importantly, these objectives differ only in how the noised input and training
target are constructed. The backbone architecture, conditioning interface,
dataset loader, and sampling code remain shared. This allows objective choices
to be studied as a controlled experimental axis rather than as separate
implementations.

\subsection{\NWM{} Architecture}

NanoWM uses a transformer backbone over latent video tokens. For VAE-style encodings, each frame is divided into spatial patches, and projected into a hidden dimension, and processed
by transformer blocks that apply interleaved spatial-temporal attention \citep{ho2022imagen,ma2024latte}. We follow the
naming convention used in image and video diffusion models: the letter denotes
the model family, while the suffix denotes the latent patch size. For example,
NanoWM-B/2 is the base model with patch size $2$, whereas NanoWM-B/4 and
NanoWM-B/8 use coarser latent patches. Nano World Models supports four architecture families: NanoWM-S, NanoWM-B,
NanoWM-L, and NanoWM-XL. These provide a scaling axis from small models for fast
iteration to larger models for higher-capacity prediction. 

\paragraph{Action Conditioning.} For action-conditioned world modeling, NanoWM conditions predictions on action
sequences $\ba_{T:T+H-1}$. The repository supports several action-injection
mechanisms. The simplest embeds actions into the transformer hidden dimension
and adds them to the corresponding frame tokens. More expressive variants inject
actions through adaptive layer normalization, fuse action and timestep
conditioning, apply FiLM-style modulation, or use cross-attention from video
tokens to action tokens. These mechanisms expose a spectrum of conditioning strategies, from lightweight
element-wise injection to higher-capacity interactions between actions and visual
dynamics.

\subsection{Latent Observation Spaces}

Following recent practice in video generation and robotic world modeling, \NWM{} predicts encoded observations rather than raw observations directly \citep{huang2025vid2world,zhou2024dino,maes2026leworldmodel}.
The choice of encoding is not merely an implementation detail: recent work
suggests that reconstruction-oriented and semantics-oriented latent spaces can
lead to different tradeoffs in visual fidelity, planning performance, and
representation quality~\citep{jha2026reconstruction}. This makes the latent
representation itself an experimental axis.

\paragraph{Supported Latent Spaces.} \NWM{} support three types of latent spaces: VAE \citep{rombach2022high}, Web-DINO \citep{fan2025scaling} and V-JEPA 2.1 \citep{mur2026v}. VAE latents provide a reconstruction-oriented space that can be decoded back
into RGB frames, making them natural for video generation and perceptual evaluation. DINO features provide a self-supervised representation space that emphasizes semantic and geometric information useful for downstream prediction
and planning. V-JEPA features provide a video-pretrained representation space designed around predictive visual features. Supporting these latent spaces under the same training interface allows \NWM{} to compare reconstruction-oriented and representation-oriented
world modeling without changing the rest of the pipeline.

\subsection{Datasets and Environment Interface}

\NWM{} uses a shared dataset and environment interface for diverse sources of sequential data. Each dataset exposes observation sequences, optional action sequences, frame windows, and metadata through the same loader
abstraction. This allows simulation environments \citep{tassa2018deepmind,fu2020d4rl}, gaming datasets \citep{pearce2022counter} and robot manipulation datasets \citep{brohan2022rt} to share the same
modeling code.

\begin{figure}[t]
    \centering
    \includegraphics[width=\linewidth]{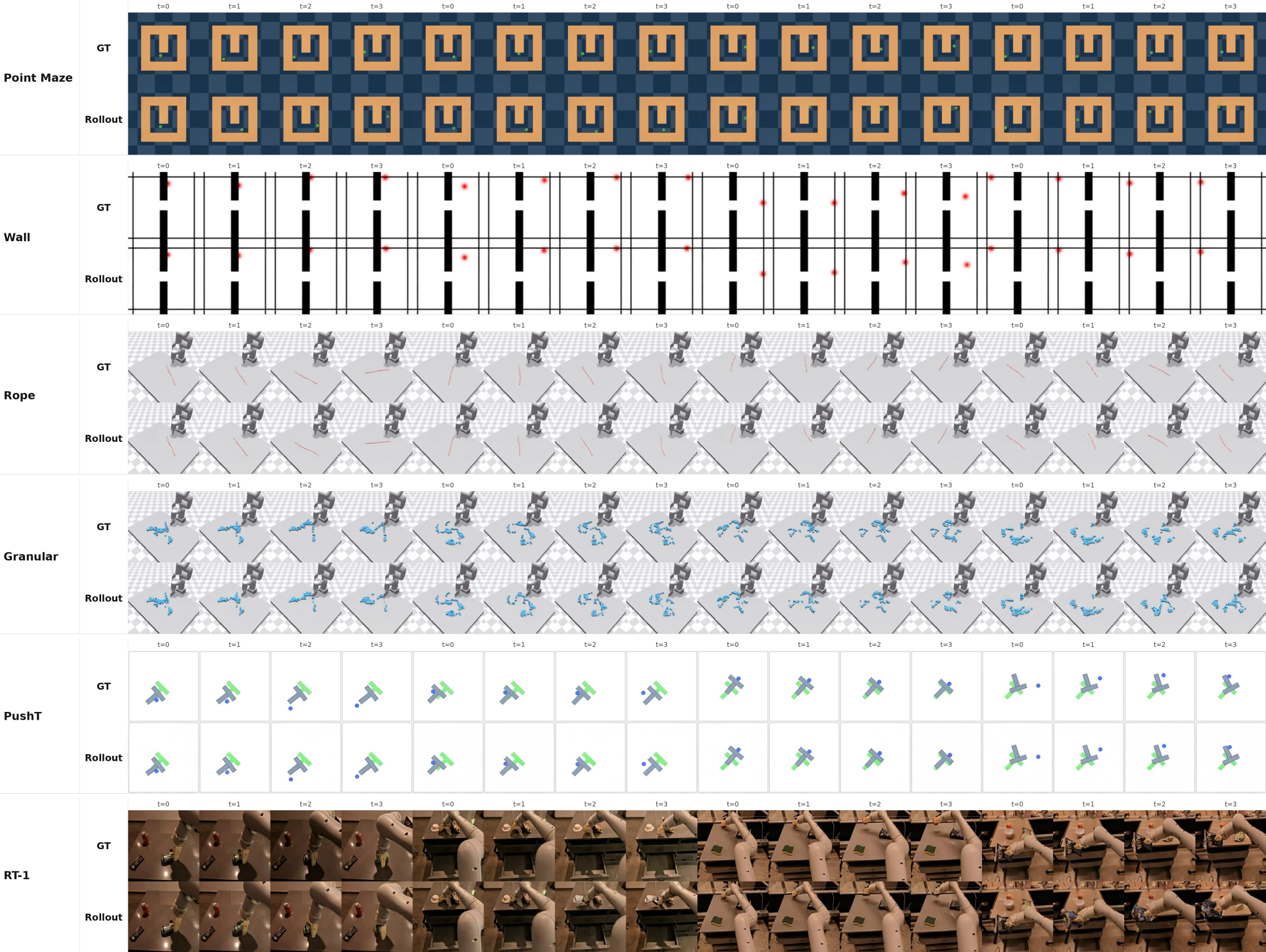}
    \caption{\textbf{Qualitative rollouts across domains.} Representative
    ground-truth (GT) sequences and \NWM{} rollouts from Point Maze, Wall, Rope,
    Granular, PushT, and RT-1. The same dataset and environment interface
    exposes these domains to the training and sampling code, allowing
    grid-world navigation, simulated control, and robot-video prediction to be
    compared under a shared rollout format.}
    \label{fig:domain-rollout-grid}
\end{figure}

\subsection{Long-Horizon Rollouts}

Although models are trained on finite windows, diffusion forcing naturally
supports generation beyond the training horizon. \NWM{} enables long-horizon generation via sliding window and auto-regressive generation: generated frames are treated as context for generation of new future frames, and the perceptive field for attention on the temporal axis follows a sliding window procedure. This enables temporally extended
rollouts while preserving the same local denoising interface used during short-horizon sampling.

\subsection{Logging, Evaluation and Reproducibility}
\begin{figure}[t]
    \centering
    \includegraphics[width=\linewidth]{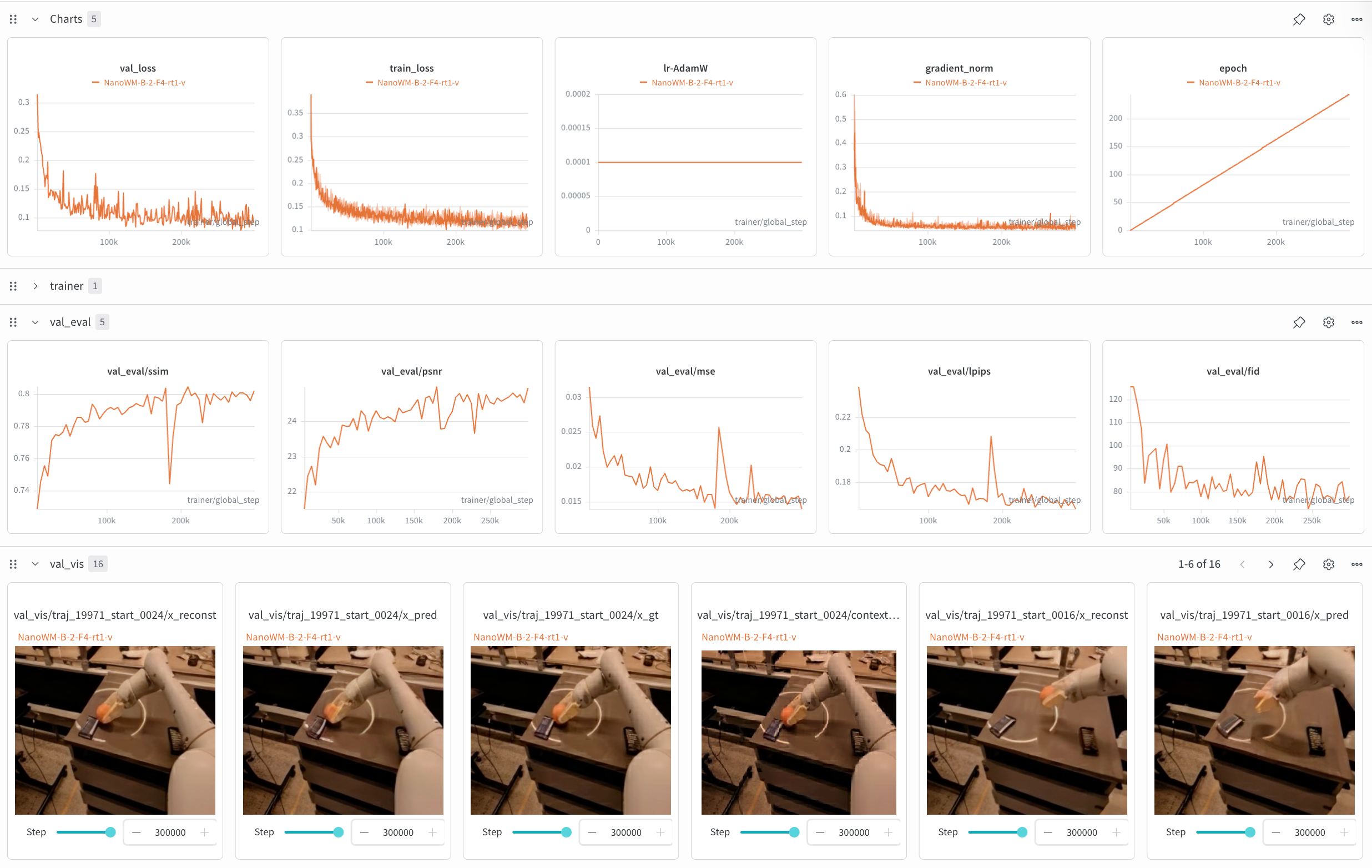}
    \caption{\textbf{Logging and fixed-seed evaluation.} \NWM{} logs training
    curves, validation metrics, and qualitative prediction panels through
    Weights \& Biases. The same callback-style evaluation pipeline records
    PSNR, SSIM, LPIPS, FID, reconstruction videos, predicted rollouts, and
    ground-truth clips under a shared run.}
    \label{fig:wandb-eval}
\end{figure}
\NWM{} includes fixed-seed validation, checkpointing, logging, and standardized evaluation scripts. For logging, we support both both Tensorboard \citep{abadi2016tensorflow} and Wandb \citep{wandb} logging systems. To further ensure reproducibility, \NWM{} open-sources all final checkpoints for supported environments, and ablated design choices.

\subsection{World-Modeling as \colorbold{Tool-Use}}
\begin{figure}[t]
    \centering
    \includegraphics[width=\linewidth]{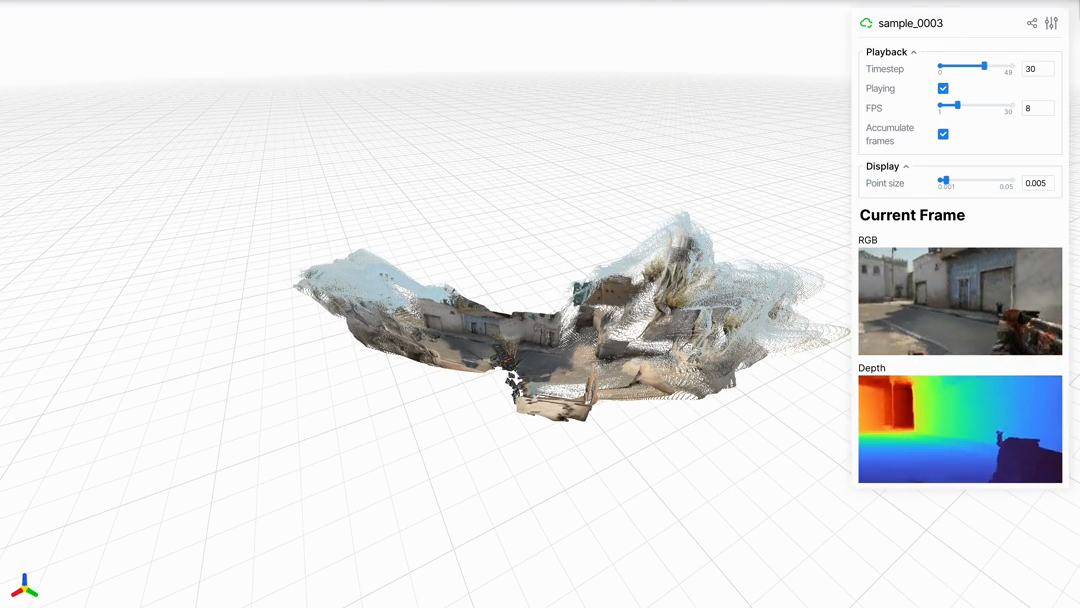}
    \caption{\textbf{Exporting rollouts to 3D scene assets.} A generated CSGO
    rollout is decoded to RGB frames and passed to a downstream depth and camera
    estimation pipeline. The resulting geometry can be visualized as a persistent
    point cloud together with the source RGB frame and estimated depth map.}
    \label{fig:video-to-3d}
\end{figure}
\paragraph{Exporting Video Rollouts to 3D Scene Assets.}
To use generated futures as inputs to 3D reconstruction tools, \NWM{} provides a
rollout export interface that saves predicted frames together with the metadata
needed by downstream reconstruction pipelines. A generated trajectory is first
decoded into RGB frames at the model's training resolution. When the source
domain has a different native aspect ratio, frames can be remapped to the native
resolution before reconstruction. The exported video or frame sequence can then
be passed to off-the-shelf multi-view depth and camera estimation systems \citep{depthanything3,chen2026geometric}, whose
outputs are converted into persistent 3D representations such as point clouds.
This design keeps the world model independent of any particular 3D backend:
\NWM{} supplies temporally coherent visual rollouts, while the reconstruction
module handles geometry estimation and visualization.

\paragraph{MPC Interface for Goal-Conditioned Planning.}
For planning, \NWM{} exposes the world model as a batched rollout function. At
each replanning step, the planner receives the current observation context, a
goal specification, and a population of candidate action sequences. The world
model predicts a future trajectory for each candidate in parallel, and an
objective module assigns a scalar score to each rollout, such as distance to a
goal state, task progress, or environment-specific reward. The planner updates
the candidate distribution, selects the best sequence, executes its first action,
and then repeats the procedure with the newly observed context. This separates
the learned dynamics model from the planning algorithm and reward definition,
allowing the same checkpoint to be reused across different goal specifications
and trajectory optimizers.


\section{Findings}

\subsection{How to measure performance?}
\paragraph{Evaluation Setup.} Unless otherwise stated, we evaluate on 256 fixed validation clips with seed 42, and auto-regressive sequential scheduling. For the standard 256-resolution models, each validation clip contains four frames: the model conditions on the first frame and generates the remaining three frames. Metrics are computed only on generated frames, excluding the context frame, while saved visualization videos include both context and generated frames. For CSGO-specific models, the
model context is four frames out of a 16-frame window; long-horizon rollouts use a sliding four-frame history window. For diffusion models, we use 250 DDIM sampling steps if not explicitly stated.

\paragraph{Measuring Visual Fidelity.} One important, though not all-encompassing factor of video world models is it's generation fidelity. We report four reconstruction and perceptual metrics. PSNR and SSIM measure pixel-level fidelity, LPIPS \citep{zhang2018unreasonable} measures perceptual distance, and FID \citep{heusel2017gans} measures
distributional similarity of generated frames to ground-truth validation frames.
For longer videos with sufficient temporal extent, FVD \citep{unterthiner2018towards} is also computed.

\paragraph{Measuring Decision-Making Capabilities.}
For goal-conditioned planning, we evaluate \NWM{} through a CEM-style MPC loop. We report decision-making performance over a fixed number of evaluation episodes, using \textit{Success Rate}, i.e. the fraction of episodes in which the final state satisfies the environment-specific goal condition as the primary metric. 

\subsection{Which objective function performs best?}

We first ablate the prediction target on RT-1 fractal \citep{brohan2022rt}. All runs use NanoWM-B/2 for model configuration, element-wise addition for action injection, and the Stable Diffusion VAE (\texttt{stabilityai/sd-vae-ft-mse}\footnote{\url{https://huggingface.co/stabilityai/sd-vae-ft-mse}}) as encoder. Each run is trained for 50K steps on 8 GPUs with per-GPU batch size 8, giving effective batch size 64. All runs condition on one frame and generate three future frames during evaluation.

We pair each prediction target with the schedule commonly used in the implementation. For $x$- and $v$-prediction, we use a squared-cosine noise schedule with zero-terminal SNR (ZTSNR \citep{lin2024common}), following the common practice of enforcing the final diffusion state to contain no residual signal. For $\beps$-prediction, we use
a linear schedule without ZTSNR, since the cosine + ZTSNR parameterization is
numerically degenerate for $\epsilon$-prediction at the terminal timestep. Thus, the comparison reflects the best supported schedule for
each target rather than forcing all targets into an unstable shared schedule.

\begin{table}[h]
\centering
\caption{Prediction target ablation on RT-1 fractal.}
\small
\begin{tabular}{lccccc}
\toprule
Target & Schedule & PSNR $\uparrow$ & SSIM $\uparrow$ & LPIPS $\downarrow$ & FID $\downarrow$ \\
\midrule
$\bv$        & cosine + ZTSNR & 23.07 & 0.760 & 0.207 & \textbf{42.27} \\
$\bx$        & cosine + ZTSNR & \textbf{23.37} & \textbf{0.783} & \textbf{0.184} & 42.99 \\
$\beps$ & linear          & 21.89 & 0.739 & 0.225 & 48.86 \\
\bottomrule
\end{tabular}
\label{tab:pred-target}
\end{table}

\begin{AIbox}{Finding \#1}
\colorbold{$\epsilon$-prediction underperforms $\bx$ and $\bv$-prediction.} $\bx$-prediction gives the best reconstruction metrics, while $\bv$-prediction gives
the best FID and is the default setting in \NWM{}. Both substantially outperform
$\beps$-prediction under the tested schedules.
\end{AIbox}

\subsection{Which architecture and action-injection?}

\paragraph{Model scale.}
We next ablate model scale on RT-1 under the same 50K-step ablation protocol: 8 GPUs, per-GPU batch size 8, effective batch size 64, one context frame, and three generated frames at evaluation time. The comparison varies the NanoWM
architecture while keeping the objective and action-injection setting fixed.

\begin{table}[h]
\centering
\caption{Model scale ablation on RT-1 fractal.}
\small
\begin{tabular}{lccccc}
\toprule
Architecture & Params & PSNR $\uparrow$ & SSIM $\uparrow$ & LPIPS $\downarrow$ & FID $\downarrow$ \\
\midrule
NanoWM-S/2 & 39.8M     & 22.30 & 0.739 & 0.230 & 54.95 \\
NanoWM-B/2 & 158.6M    & 23.07 & 0.760 & 0.207 & 42.27 \\
NanoWM-L/2 & $\sim$460M & \textbf{23.62} & \textbf{0.777} & \textbf{0.186} & \textbf{36.31} \\
\bottomrule
\end{tabular}
\label{tab:model-scale}
\end{table}

\begin{AIbox}{Finding \#2}
\colorbold{Larger Models give Better Performance.} Scaling from NanoWM-S/2 to NanoWM-B/2 to NanoWM-L/2 improves PSNR, SSIM, LPIPS,
and FID monotonically on RT-1.
\end{AIbox}

\paragraph{Action injection.}
We compare five action-injection mechanisms: Element-wise Addition (\textit{additive}), Adaptive Layer Norm (\textit{adaLN}), Adaptive Layer Norm fused with timestep injection (\textit{adaLN-fuse}), FiLM (\textit{FiLM}) and cross-attention (\textit{cross-attention}). On RT-1, each run uses the same 50K-step ablation protocol as above. We also report a PushT sweep with NanoWM-B/2 trained for 30K steps and evaluated on 256 fixed validation samples
with seed 42.

\begin{table}[h]
\centering
\caption{Action-injection ablations on RT-1 and PushT.}
\scriptsize
\begin{minipage}{0.49\linewidth}
\centering
\resizebox{\linewidth}{!}{
\begin{tabular}{lccccc}
\toprule
RT-1 method & PSNR & SSIM & LPIPS & FID & Params \\
\midrule
additive        & 23.07 & 0.760 & 0.207 & 42.27 & 158.6M \\
adaLN           & 23.19 & 0.762 & 0.206 & 43.62 & 158.6M \\
adaLN-fuse      & 23.10 & 0.762 & 0.206 & 43.03 & 158.6M \\
FiLM            & \textbf{23.20} & \textbf{0.763} & \textbf{0.203} & \textbf{40.62} & 172.8M \\
cross-attention & 20.82 & 0.721 & 0.242 & 51.12 & 187.0M \\
\bottomrule
\end{tabular}}
\end{minipage}
\hfill
\begin{minipage}{0.49\linewidth}
\centering
\resizebox{\linewidth}{!}{
\begin{tabular}{lccccc}
\toprule
PushT method & PSNR & SSIM & LPIPS & FID & Extra params \\
\midrule
additive        & \textbf{26.20} & \textbf{0.962} & 0.053 & \textbf{23.89} & 0 \\
adaLN-fuse      & 26.17 & 0.961 & \textbf{0.051} & 30.28 & 0 \\
adaLN           & 26.09 & 0.960 & 0.053 & 26.32 & $\sim$42.5M \\
cross-attention & 25.95 & 0.959 & 0.055 & 28.64 & $\sim$28.3M \\
FiLM            & 25.88 & 0.960 & 0.056 & 25.45 & $\sim$14.4M \\
\bottomrule
\end{tabular}}
\end{minipage}
\label{tab:action-injection}
\end{table}

\begin{AIbox}{Finding \#3}
\colorbold{Action injection is task-dependent}. FiLM gives the best visual fidelity on RT-1, but the simple additive baseline is strongest on PushT and has the best quality-parameter tradeoff.
\end{AIbox}

\subsection{Which latent space?}

\NWM{} supports multiple latent observation spaces, including reconstruction-oriented VAE latents and semantic representation latents such as Web-DINO \citep{fan2025scaling} and V-JEPA 2.1 \citep{mur2026v}. Unlike VAE latents, Web-DINO and V-JEPA features are not naturally decoded back to RGB observations, and therefore visual fidelity metrics such as PSNR, SSIM, LPIPS, and FID are not directly comparable across latent spaces. We instead evaluate whether each latent space can support model-based control through goal-conditioned planning.

\paragraph{Environment Setup.} We evaluate on PushT goal reaching using the DINO-WM PushT dataset with a frame interval of 5 and
a four-frame prediction window consisting of one context frame and three future frames. Actions are
represented as relative actions, and
flattened from a 5-step chunk of 2D actions into a 10-dimensional action vector. All models are trained
for 100K steps with diffusion forcing, $v$-prediction, a squared-cosine noise schedule with zero-terminal SNR, AdamW with learning rate $10^{-4}$ and weight decay $0.01$, causal masking, and additive action injection.

\paragraph{Model Setup.} For fair comparison with respect to token numbers, we compare three checkpoints with the following configuration. The VAE model uses Stable Diffusion VAE latents with shape $[4,32,32]$ and a NanoWM-B/2 backbone. The Web-DINO and V-JEPA 2.1 models use dense semantic latents with shape $[1024,16,16]$ and NanoWM-B/1 backbones. Web-DINO is applied frame-wise with 224-resolution inputs and patch size 14, while V-JEPA 2.1 uses the EMA encoder, with the input resolution set to 256 to align the latent
grid and patch size 16.

\paragraph{Evaluation.} For planning, we use a CEM-style MPC loop. At each replanning step, the planner samples candidate
action sequences, rolls out the world model for one autoregressive chunk, and scores the final predicted
latent frame against a replayed dataset goal. We use goal horizon $H=3$, so the model generates three
future frames and the last predicted frame is matched to the goal. We evaluate each model using 64 CEM samples and 5 CEM iterations.

\begin{table}[t]
\centering
\caption{
\textbf{Goal-conditioned planning on PushT across latent spaces.}
Success is measured by the environment's state-based PushT success criterion. Although all models
are trained under the same diffusion-forcing interface, only the reconstruction-oriented VAE latent
checkpoint produces non-zero planning success.
}
\label{tab:latent_planning}
\begin{tabular}{l c c c}
\toprule
\textbf{Latent space} & \textbf{Backbone} & \textbf{Latent shape} & \textbf{Success rate} \\
\midrule
SD-VAE & NanoWM-B/2 & $[4,32,32]$ & $25.0\%$ \\
Web-DINO & NanoWM-B/1 & $[1024,16,16]$ & $0.0\%$ \\
V-JEPA 2.1 & NanoWM-B/1 & $[1024,16,16]$ & $0.0\%$ \\
\bottomrule
\end{tabular}
\end{table}

\paragraph{Results.} As shown in Table~\ref{tab:latent_planning}, the SD-VAE checkpoint reaches the goal in 25\% of
episodes, while both Web-DINO and V-JEPA 2.1 fail to obtain non-zero success. Internally, we also tried increasing sampling and planning budget, yet semantic-latent checkpoints still yields zero successful episodes. This suggests that
the failure is not primarily caused by an insufficient trajectory optimizer or a weak sampling budget.

\begin{table}[t]
\centering
\caption{
\textbf{Ground-truth action rollout diagnostic on PushT.}
We report the distance between the final predicted latent and the goal latent over 32 goal-reaching
episodes, using goal horizon $H=3$ and 20 DDIM sampling steps. A controllable dynamics model should
make ground-truth actions substantially closer to the goal than zero or random actions. SD-VAE shows
a clear action-conditioned improvement, while Web-DINO and V-JEPA 2.1 remain nearly unchanged
across ground-truth, zero, and random actions.
}
\label{tab:gt_action_rollout}
\resizebox{\linewidth}{!}{
\begin{tabular}{l cccc cccc}
\toprule
\multirow{2}{*}{\textbf{Latent space}}
& \multicolumn{4}{c}{\textbf{Latent MSE} $\downarrow$}
& \multicolumn{4}{c}{\textbf{Cosine distance} $\downarrow$} \\
\cmidrule(lr){2-5} \cmidrule(lr){6-9}
& \textbf{Init} & \textbf{GT action} & \textbf{Zero action} & \textbf{Random action}
& \textbf{Init} & \textbf{GT action} & \textbf{Zero action} & \textbf{Random action} \\
\midrule
SD-VAE
& 0.077714 & \textbf{0.014015} & 0.074830 & 0.081412
& 0.038073 & \textbf{0.008885} & 0.037322 & 0.042239 \\
Web-DINO
& 0.311649 & 0.834037 & 0.834044 & 0.834066
& 0.111740 & 0.280007 & 0.280011 & 0.280025 \\
V-JEPA 2.1
& 0.206433 & 0.584029 & 0.584056 & 0.584150
& 0.047607 & 0.138866 & 0.138872 & 0.138893 \\
\bottomrule
\end{tabular}
}
\end{table}

\begin{table}[h]
\centering
\caption{
\textbf{Action embedding magnitude across latent spaces.}
The semantic-latent checkpoints learn action embeddings with near-zero RMS, indicating that the
additive action-conditioning pathway is effectively unused.
}
\label{tab:action_rms}
\begin{tabular}{l c}
\toprule
\textbf{Latent space} & \textbf{Action embedding RMS} \\
\midrule
SD-VAE & 0.1119 \\
Web-DINO & 0.00214 \\
V-JEPA 2.1 & 0.00129 \\
\bottomrule
\end{tabular}
\end{table}

\paragraph{Analysis.} To understand why semantic-latent planning fails, we test whether the trained dynamics models
actually use the action input. We first compare rollouts under ground-truth actions against rollouts under zero or random actions. For each evaluation goal, we measure the distance between the predicted final latent and the goal
latent after a 3-frame rollout with 20 DDIM steps. We also compare against the initial observation
latent, which measures whether the model predicts progress toward the goal beyond simply staying
near the current state. 

As shown in Table~\ref{tab:gt_action_rollout}, the SD-VAE checkpoint is strongly sensitive to
ground-truth actions; in contrast, Web-DINO and V-JEPA 2.1 show almost identical distances under ground-truth, zero, and random actions. This indicates that the
semantic-latent checkpoints do not meaningfully condition their predictions on the action input. The same conclusion is supported by the magnitude of the learned action branch, as shown in Table \ref{tab:action_rms}. The action embedding
Root Mean Square (RMS) is much larger for SD-VAE compared to Web-DINO and V-JEPA 2.1. Together, these results indicate that the reason behind semantic encoders' incompetence in aiding planning comes from inability to learn counterfactual action-conditioned predictions. 

\begin{AIbox}{Finding \#4}
\colorbold{Semantic latent spaces do not automatically yield better world models for planning.} On PushT, the SD-VAE
checkpoint learns action-conditioned dynamics and achieves non-zero goal-reaching success, whereas
the Web-DINO and V-JEPA 2.1 checkpoints become nearly action-agnostic under the same training
interface and fails completely. Their planning failure is caused by the semantic-latent diffusion objective not sufficiently forcing the model to use actions for controllable dynamics.
\end{AIbox}

\subsection{What happens during long-horizon rollouts?}

\begin{figure}[t]
        \centering
        \includegraphics[width=\linewidth]{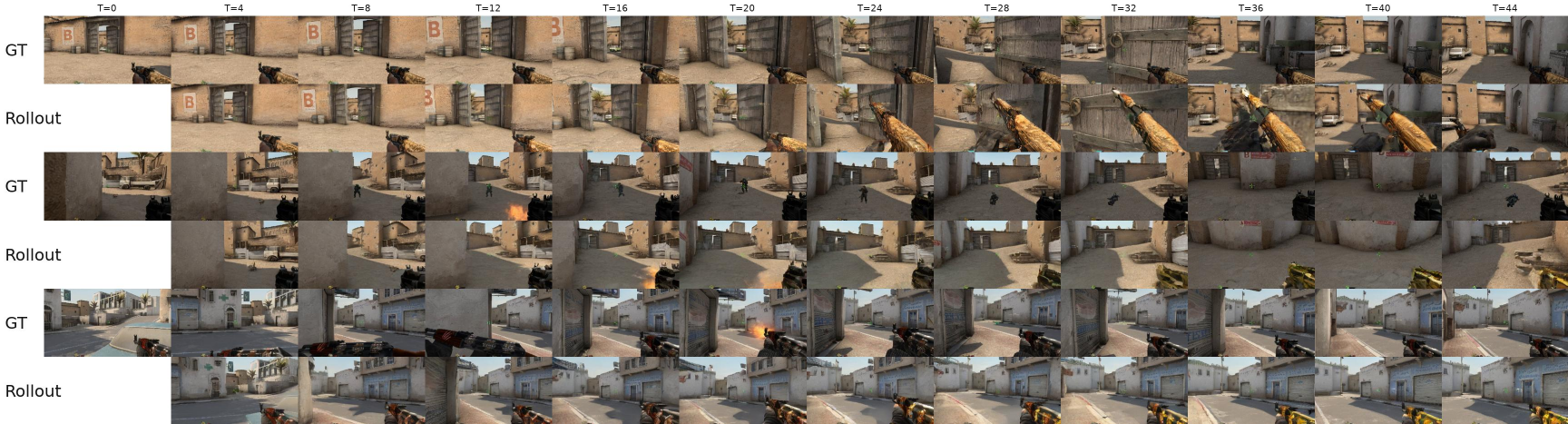}
        \captionof{figure}{Long-Horizon Rollout Samples of \NWM{}.}
        \label{fig:long_rollout_samples}
\end{figure}

\begin{wrapfigure}{r}{0.40\linewidth}
    \vspace{-0.8em}
    \centering
    \includegraphics[width=\linewidth]{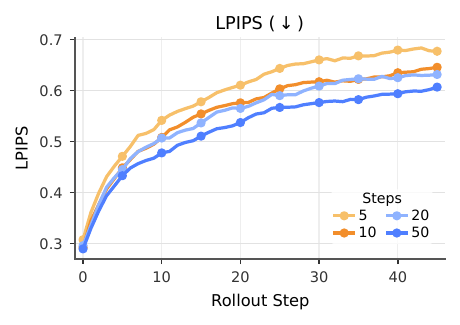}
    \caption{Error Accumulation.}
    \label{fig:timeline_lpips}
    \vspace{-1.0em}
\end{wrapfigure}

For long-horizon qualitative evaluation, the CSGO checkpoint uses the
CSGO-specific NanoWM-L/2 configuration with 16-frame training windows and
four context frames. The long-rollout script generates 50-frame videos by
initializing from four ground-truth history frames and then autoregressively
generating the remaining 46 frames one at a time. Each generation step uses a
sliding four-frame context window, sequential scheduling, and 50 DDIM sampling
steps per predicted frame.

As shown in Fig.~\ref{fig:long_rollout_samples}, \NWM{} preserves the coarse scene geometry and camera motion over long autoregressive rollouts, while gradually accumulating perceptual errors in fine-grained visual details such as weapon appearance and local textures. This behavior is consistent with the LPIPS curves in Fig.~\ref{fig:timeline_lpips}: prediction error increases gradually during self-generated rollouts. Increasing the number of DDIM sampling steps consistently reduces LPIPS across the rollout horizon, suggesting that more accurate per-frame denoising mitigates compounding errors during autoregressive generation.

\begin{AIbox}{Finding \#5}
\NWM{} can produce plausible long-horizon rollouts, but autoregressive generation inevitably accumulates perceptual errors over time. Increasing the DDIM sampling budget consistently improves rollout fidelity, suggesting that stronger per-frame denoising can partially mitigate compounding errors.
\end{AIbox}

\subsection{How do findings vary across task and environment?}

Finally, we evaluate the shipped checkpoints across domains. DINO-WM Point Maze,
Wall, Rope, and Granular use 2 GPUs with per-GPU batch size 8. DINO-WM PushT
uses 8 GPUs with per-GPU batch size 8. RT-1 uses 8 GPUs with per-GPU batch size
8 and trains for 300K steps. All rows below use the standardized evaluation
protocol: 256 fixed validation clips with seed 42, 250 DDIM sampling steps,
sequential scheduling, one context frame, and three generated frames.

\begin{table}[h]
\centering
\caption{Results on shipped checkpoints under the standardized evaluation
protocol.}
\small
\begin{tabular}{lccccc}
\toprule
Dataset & Steps & PSNR $\uparrow$ & SSIM $\uparrow$ & LPIPS $\downarrow$ & FID $\downarrow$ \\
\midrule
Point Maze & 30K  & \textbf{36.74} & 0.984 & 0.019 & 9.66 \\
Wall       & 15K  & 34.05 & \textbf{0.994} & \textbf{0.010} & \textbf{2.64} \\
Rope       & 15K  & 31.63 & 0.953 & 0.056 & 35.20 \\
Granular   & 15K  & 26.08 & 0.917 & 0.073 & 40.05 \\
PushT      & 100K & 33.19 & 0.982 & 0.016 & 13.63 \\
RT-1       & 300K & 24.36 & 0.787 & 0.180 & 35.08 \\
\bottomrule
\end{tabular}
\label{tab:shipped-checkpoints}
\end{table}

\begin{AIbox}{Finding \#6}
The same training and evaluation recipe works across navigation, tabletop
pushing, deformable manipulation, and real-robot data. Performance is strongest
on simpler simulated domains and decreases on visually or dynamically more
complex domains such as Granular and RT-1.
\end{AIbox}

\newcommand{\atsign}{\makeatletter @ \makeatother }
\section{Related Works}

\paragraph{Representative Modern World-Modeling Paradigms.}
World models aim to learn predictive representations of the environment that can support generation, planning, and decision-making. One line of works \citep{ha2018recurrent,dawid2023introduction} focuses on learning representation spaces that facilitate decision-making, with the notable example of the JEPA family of models \citep{bardes2024revisiting,balestriero2025lejepaprovablescalableselfsupervised,mur2026v,maes2026leworldmodel}, which jointly learns encoder and predictor in a self-supervised fashion. A second line of work focuses on generations in 3D space, using 3D structures as representation. While they start from static 3D asset generation \citep{worldlabs2025marble,shen2026lyra}, recent works have included temporal evolutions and next state predictions given interactive actions \citep{zhen2025tesseract,huang2026pointworld}. The thrid line of work leverages video generation models \citep{huang2025vid2world}, focusing on generating plausible future observations in the form of RGB images. This line of work have shown promising results in game simulation (i.e. \textit{neural game engines}) \citep{alonso2024diffusion,hafner2025training,savva2026solaris}, aiding navigation \citep{bar2025navigation} as well as robot manipulation \citep{guo2025ctrl,chi2025wow}.\NWM{} primarily follows this video-generative view, but supports the latent observation space as a design axis.

\paragraph{Streaming Video Diffusion.}
Standard video diffusion models \citep{ho2022video,svd} are usually trained to generate fixed-length clips in a full-sequence manner, which makes them poorly matched to online interaction, long-horizon rollout, and real-time control. Recent work on streaming and autoregressive video diffusion addresses this limitation by generating videos progressively \citep{chen2024diffusion,kodaira2025streamdiffusion,yang2025longlive,huang2026self}. These models highlight a central challenge for video world modeling: a finite-window generator must be converted into a persistent simulator without losing temporal consistency or accumulating excessive visual drift. \NWM{} studies this problem in a compact and controllable setting. Through diffusion forcing, sequential sampling schedules, and sliding-window autoregressive rollouts, NanoWM uses a single diffusion-based interface for short-horizon prediction and extended future generation, while making the effects of sampling budget and compounding error directly measurable.

\paragraph{Advancing Open-Source World Modeling.}
Several recent efforts have begun to close the gap between closed, industry-scale world simulators and reproducible academic research infrastructure. \textit{StableWM} \citep{maes_lelidec2026swm} provides an open platform for world-model research, covering data collection, training, evaluation, and planning-oriented workflows, however mostly limited to simple simulation environment. \textit{Jasmine} \citep{mahajan2025jasmine} emphasizes highly efficient world-model training infrastructure, based on JAX \citep{jax2018github} implementations, however supporting mostly outdated algorithms such as Genie \citep{bruce2024genie}. \textit{LingBot-World} \citep{lingbot-world} advances open-source interactive world simulation from video generation, releasing industry-scale models and code for long-horizon, real-time, action-controllable environments, yet posing forbidding costs to train or even inference. These efforts share the goal of democratizing world-model research, but differ in scope and emphasis. \NWM{} is complementary: rather than targeting the largest possible simulator, it provides a minimalist PyTorch implementation centered on diffusion-forcing future video prediction, with modular support for objectives, architecture sizes, action-injection mechanisms, latent spaces, datasets, evaluations, long-horizon rollout, and released checkpoints, aiming to make careful and scientific understanding of design choices possible.
\section{Conclusion}

We introduce \NWM{}, a minimalist and reproducible framework for advancing scientific understanding of world models. \NWM{} treats diffusion forcing \citep{chen2024diffusion} as a unifying abstraction, allowing objectives, architectures, action-conditioning mechanisms, latent spaces, datasets, and rollout procedures to be varied within a shared training and evaluation pipeline. Our empirical results show that these design choices have measurable and often domain-dependent effects: prediction parameterization changes reconstruction and distributional quality, scaling improves performance consistently, action injection interacts with task structure, and long-horizon autoregressive rollout remains limited by compounding perceptual error. Together, these findings highlight the need for world-model research to move beyond \textit{isolated demonstrations} toward \textit{controlled studies of the modeling decisions that govern prediction, interaction, and planning}. \NWM{} provides an open substrate for this direction, making it easier to compare design choices, reproduce results, and build future world models on common experimental ground.

\newpage
\newpage
\bibliographystyle{plainnat}
\bibliography{refs}

@inproceedings{perez2018film,
  title={Film: Visual reasoning with a general conditioning layer},
  author={Perez, Ethan and Strub, Florian and De Vries, Harm and Dumoulin, Vincent and Courville, Aaron},
  booktitle={Proceedings of the AAAI conference on artificial intelligence},
  volume={32},
  number={1},
  year={2018}
}

@article{ha2018recurrent,
  title={Recurrent world models facilitate policy evolution},
  author={Ha, David and Schmidhuber, J{\"u}rgen},
  journal={Advances in neural information processing systems},
  volume={31},
  year={2018}
}

@article{dawid2023introduction,
  title     = {Introduction to Latent Variable Energy-Based Models: A Path Towards Autonomous Machine Intelligence},
  author    = {Anna Dawid and Yann LeCun},
  journal   = {Journal of Statistical Mechanics: Theory and Experiment},
  year      = {2023},
  doi       = {10.1088/1742-5468/ad292b},
}

@article{richens2025general,
  title={General agents contain world models},
  author={Richens, Jonathan and Abel, David and Bellot, Alexis and Everitt, Tom},
  journal={arXiv preprint arXiv:2506.01622},
  year={2025}
}

@article{guo2025ctrl,
  title={Ctrl-world: A controllable generative world model for robot manipulation},
  author={Guo, Yanjiang and Shi, Lucy Xiaoyang and Chen, Jianyu and Finn, Chelsea},
  journal={arXiv preprint arXiv:2510.10125},
  year={2025}
}

@inproceedings{yang2025cambrian,
  title={Cambrian-s: Towards spatial supersensing in video},
  author={Yang, Shusheng and Yang, Jihan and Huang, Pinzhi and Brown II, Ellis L and Yang, Zihao and Yu, Yue and Tong, Shengbang and Zheng, Zihan and Xu, Yifan and Wang, Muhan and others},
  booktitle={The Fourteenth International Conference on Learning Representations},
  year={2025}
}

@article{wang2026vagen,
  title={Vagen: Reinforcing world model reasoning for multi-turn vlm agents},
  author={Wang, Kangrui and Zhang, Pingyue and Wang, Zihan and Gao, Yaning and Li, Linjie and Wang, Qineng and Chen, Hanyang and Lu, Yiping and Yang, Zhengyuan and Wang, Lijuan and others},
  journal={Advances in Neural Information Processing Systems},
  volume={38},
  pages={172871--172933},
  year={2026}
}

@misc{parkerholder2025genie3,
  author       = {Google},
  title        = {Genie 3: A new frontier for world models},
  year         = {2025},
  month        = {August},
  howpublished = {\url{https://deepmind.google/blog/genie-3-a-new-frontier-for-world-models/}},
  note         = {Google DeepMind Blog},
}

@article{lingbot-world,
      title={Advancing Open-source World Models}, 
      author={Robbyant Team and Zelin Gao and Qiuyu Wang and Yanhong Zeng and Jiapeng Zhu and Ka Leong Cheng and Yixuan Li and Hanlin Wang and Yinghao Xu and Shuailei Ma and Yihang Chen and Jie Liu and Yansong Cheng and Yao Yao and Jiayi Zhu and Yihao Meng and Kecheng Zheng and Qingyan Bai and Jingye Chen and Zehong Shen and Yue Yu and Xing Zhu and Yujun Shen and Hao Ouyang},
      journal={arXiv preprint arXiv:2601.20540},
      year={2026}
}

@article{ho2022video,
  title={Video diffusion models},
  author={Ho, Jonathan and Salimans, Tim and Gritsenko, Alexey and Chan, William and Norouzi, Mohammad and Fleet, David J},
  journal={Advances in neural information processing systems},
  volume={35},
  pages={8633--8646},
  year={2022}
}

@article{svd,
  title   = {Stable Video Diffusion: Scaling Latent Video Diffusion Models to Large Datasets},
  author  = {Andreas Blattmann and Tim Dockhorn and Sumith Kulal and Daniel Mendelevitch and Maciej Kilian and Dominik Lorenz and Yam Levi and Zion English and Vikram Voleti and Adam Letts and Varun Jampani and Robin Rombach},
  year    = {2023},
  journal = {arXiv preprint arXiv: 2311.15127}
}

@article{chen2024diffusion,
  title={Diffusion forcing: Next-token prediction meets full-sequence diffusion},
  author={Chen, Boyuan and Mart{\'\i} Mons{\'o}, Diego and Du, Yilun and Simchowitz, Max and Tedrake, Russ and Sitzmann, Vincent},
  journal={Advances in Neural Information Processing Systems},
  volume={37},
  pages={24081--24125},
  year={2024}
}

@article{song2023consistency,
  title={Consistency models},
  author={Song, Yang and Dhariwal, Prafulla and Chen, Mark and Sutskever, Ilya},
  year={2023}
}

@article{zhou2024dino,
  title={Dino-wm: World models on pre-trained visual features enable zero-shot planning},
  author={Zhou, Gaoyue and Pan, Hengkai and LeCun, Yann and Pinto, Lerrel},
  journal={arXiv preprint arXiv:2411.04983},
  year={2024}
}

@article{brohan2022rt,
  title={Rt-1: Robotics transformer for real-world control at scale},
  author={Brohan, Anthony and Brown, Noah and Carbajal, Justice and Chebotar, Yevgen and Dabis, Joseph and Finn, Chelsea and Gopalakrishnan, Keerthana and Hausman, Karol and Herzog, Alex and Hsu, Jasmine and others},
  journal={arXiv preprint arXiv:2212.06817},
  year={2022}
}

@inproceedings{pearce2022counter,
  title={Counter-strike deathmatch with large-scale behavioural cloning},
  author={Pearce, Tim and Zhu, Jun},
  booktitle={2022 IEEE Conference on Games (CoG)},
  pages={104--111},
  year={2022},
  organization={IEEE}
}

@article{li2025back,
  title={Back to basics: Let denoising generative models denoise},
  author={Li, Tianhong and He, Kaiming},
  journal={arXiv preprint arXiv:2511.13720},
  year={2025}
}

@article{lipman2024flow,
  title={Flow matching guide and code},
  author={Lipman, Yaron and Havasi, Marton and Holderrieth, Peter and Shaul, Neta and Le, Matt and Karrer, Brian and Chen, Ricky TQ and Lopez-Paz, David and Ben-Hamu, Heli and Gat, Itai},
  journal={arXiv preprint arXiv:2412.06264},
  year={2024}
}

@article{yang2023diffusion,
  title={Diffusion models: A comprehensive survey of methods and applications},
  author={Yang, Ling and Zhang, Zhilong and Song, Yang and Hong, Shenda and Xu, Runsheng and Zhao, Yue and Zhang, Wentao and Cui, Bin and Yang, Ming-Hsuan},
  journal={ACM computing surveys},
  volume={56},
  number={4},
  pages={1--39},
  year={2023},
  publisher={ACM New York, NY, USA}
}

@inproceedings{
quevedo2026worldgym,
title={WorldGym: World Model as An Environment for Policy Evaluation},
author={Julian Hector Quevedo and Ansh Kumar Sharma and Yixiang Sun and Varad Suryavanshi and Percy Liang and Sherry Yang},
booktitle={The Fourteenth International Conference on Learning Representations},
year={2026},
url={https://openreview.net/forum?id=hidBHy1CAw}
}

@article{alonso2024diffusion,
  title={Diffusion for world modeling: Visual details matter in atari},
  author={Alonso, Eloi and Jelley, Adam and Micheli, Vincent and Kanervisto, Anssi and Storkey, Amos and Pearce, Tim and Fleuret, Fran{\c{c}}ois},
  journal={Advances in Neural Information Processing Systems},
  volume={37},
  pages={58757--58791},
  year={2024}
}

@inproceedings{vafa2024world,
  title={Evaluating the World Model Implicit in a Generative Model},
  author={Vafa, Keyon and Chen, Justin Y and Rambachan, Ashesh and Kleinberg, Jon and Mullainathan, Sendhil},
  booktitle={Neural Information Processing Systems},
  year={2024},
}

@inproceedings{
zhang2026worldinworld,
title={World-In-World: World Models in a Closed-Loop World},
author={Jiahan Zhang and Muqing Jiang and Nanru Dai and TaiMing Lu and Arda Uzunoglu and Shunchi Zhang and Yana Wei and Jiahao Wang and Vishal M. Patel and Paul Pu Liang and Daniel Khashabi and Cheng Peng and Rama Chellappa and Tianmin Shu and Alan Yuille and Yilun Du and Jieneng Chen},
booktitle={The Fourteenth International Conference on Learning Representations},
year={2026},
url={https://openreview.net/forum?id=yDmb7xAfeb}
}

@Misc{Yadan2019Hydra,
  author =       {Omry Yadan},
  title =        {Hydra - A framework for elegantly configuring complex applications},
  howpublished = {Github},
  year =         {2019},
  url =          {https://github.com/facebookresearch/hydra}
}

@inproceedings{peebles2023scalable,
  title={Scalable Diffusion Models with Transformers},
  author={Peebles, William and Xie, Saining},
  booktitle={2023 IEEE/CVF International Conference on Computer Vision (ICCV)},
  pages={4172--4182},
  year={2023},
  organization={IEEE}
}

@inproceedings{rombach2022high,
  title={High-resolution image synthesis with latent diffusion models},
  author={Rombach, Robin and Blattmann, Andreas and Lorenz, Dominik and Esser, Patrick and Ommer, Bj{\"o}rn},
  booktitle={Proceedings of the IEEE/CVF conference on computer vision and pattern recognition},
  pages={10684--10695},
  year={2022}
}

@article{lipman2022flow,
  title={Flow matching for generative modeling},
  author={Lipman, Yaron and Chen, Ricky TQ and Ben-Hamu, Heli and Nickel, Maximilian and Le, Matt},
  journal={arXiv preprint arXiv:2210.02747},
  year={2022}
}

@article{tassa2018deepmind,
  title={Deepmind control suite},
  author={Tassa, Yuval and Doron, Yotam and Muldal, Alistair and Erez, Tom and Li, Yazhe and Casas, Diego de Las and Budden, David and Abdolmaleki, Abbas and Merel, Josh and Lefrancq, Andrew and others},
  journal={arXiv preprint arXiv:1801.00690},
  year={2018}
}

@article{fu2020d4rl,
  title={D4rl: Datasets for deep data-driven reinforcement learning},
  author={Fu, Justin and Kumar, Aviral and Nachum, Ofir and Tucker, George and Levine, Sergey},
  journal={arXiv preprint arXiv:2004.07219},
  year={2020}
}

@inproceedings{ma2024sit,
  title={Sit: Exploring flow and diffusion-based generative models with scalable interpolant transformers},
  author={Ma, Nanye and Goldstein, Mark and Albergo, Michael S and Boffi, Nicholas M and Vanden-Eijnden, Eric and Xie, Saining},
  booktitle={European Conference on Computer Vision},
  pages={23--40},
  year={2024},
  organization={Springer}
}

@inproceedings{huang2017arbitrary,
  title={Arbitrary style transfer in real-time with adaptive instance normalization},
  author={Huang, Xun and Belongie, Serge},
  booktitle={Proceedings of the IEEE international conference on computer vision},
  pages={1501--1510},
  year={2017}
}

@misc{wandb,
title = {Experiment Tracking with Weights and Biases},
year = {2020},
note = {Software available from wandb.com},
url={https://www.wandb.com/},
author = {Biewald, Lukas},
}

@article{abadi2016tensorflow,
  title={Tensorflow: Large-scale machine learning on heterogeneous distributed systems},
  author={Abadi, Mart{\'\i}n and Agarwal, Ashish and Barham, Paul and Brevdo, Eugene and Chen, Zhifeng and Citro, Craig and Corrado, Greg S and Davis, Andy and Dean, Jeffrey and Devin, Matthieu and others},
  journal={arXiv preprint arXiv:1603.04467},
  year={2016}
}

@article{depthanything3,
  title={Depth Anything 3: Recovering the visual space from any views},
  author={Haotong Lin and Sili Chen and Jun Hao Liew and Donny Y. Chen and Zhenyu Li and Guang Shi and Jiashi Feng and Bingyi Kang},
  journal={arXiv preprint arXiv:2511.10647},
  year={2025}
}

@article{chen2026geometric,
  title={Geometric Context Transformer for Streaming 3D Reconstruction},
  author={Chen, Lin-Zhuo and Gao, Jian and Chen, Yihang and Cheng, Ka Leong and Sun, Yipengjing and Hu, Liangxiao and Xue, Nan and Zhu, Xing and Shen, Yujun and Yao, Yao and Xu, Yinghao},
  journal={arXiv preprint arXiv:2604.14141},
  year={2026}
}

@article{kingma2013auto,
  title={Auto-encoding variational bayes},
  author={Kingma, Diederik P and Welling, Max},
  journal={arXiv preprint arXiv:1312.6114},
  year={2013}
}

@inproceedings{caron2021emerging,
  title={Emerging properties in self-supervised vision transformers},
  author={Caron, Mathilde and Touvron, Hugo and Misra, Ishan and J{\'e}gou, Herv{\'e} and Mairal, Julien and Bojanowski, Piotr and Joulin, Armand},
  booktitle={Proceedings of the IEEE/CVF international conference on computer vision},
  pages={9650--9660},
  year={2021}
}

@article{bardes2024revisiting,
  title={Revisiting Feature Prediction for Learning Visual Representations from Video},
  author={Bardes, Adrien and Garrido, Quentin and Ponce, Jean and Rabbat, Michael and LeCun, Yann and Assran, Mahmoud and Ballas, Nicolas},
  journal={arXiv:2404.08471},
  year={2024}
}

@article{ho2022imagen,
  title={Imagen video: High definition video generation with diffusion models},
  author={Ho, Jonathan and Chan, William and Saharia, Chitwan and Whang, Jay and Gao, Ruiqi and Gritsenko, Alexey and Kingma, Diederik P and Poole, Ben and Norouzi, Mohammad and Fleet, David J and others},
  journal={arXiv preprint arXiv:2210.02303},
  year={2022}
}

@article{ma2024latte,
  title={Latte: Latent diffusion transformer for video generation},
  author={Ma, Xin and Wang, Yaohui and Chen, Xinyuan and Jia, Gengyun and Liu, Ziwei and Li, Yuan-Fang and Chen, Cunjian and Qiao, Yu},
  journal={arXiv preprint arXiv:2401.03048},
  year={2024}
}

@article{huang2025vid2world,
  title={Vid2world: Crafting video diffusion models to interactive world models},
  author={Huang, Siqiao and Wu, Jialong and Zhou, Qixing and Miao, Shangchen and Long, Mingsheng},
  journal={arXiv preprint arXiv:2505.14357},
  year={2025}
}

@article{maes2026leworldmodel,
  title={Leworldmodel: Stable end-to-end joint-embedding predictive architecture from pixels},
  author={Maes, Lucas and Lidec, Quentin Le and Scieur, Damien and LeCun, Yann and Balestriero, Randall},
  journal={arXiv preprint arXiv:2603.19312},
  year={2026}
}

@article{jha2026reconstruction,
  title={Reconstruction or Semantics? What Makes a Latent Space Useful for Robotic World Models},
  author={Jha, Saurav and Zholus, Artem and Chandar, Sarath and others},
  journal={arXiv preprint arXiv:2605.06388},
  year={2026}
}

@inproceedings{fan2025scaling,
  title={Scaling language-free visual representation learning},
  author={Fan, David and Tong, Shengbang and Zhu, Jiachen and Sinha, Koustuv and Liu, Zhuang and Chen, Xinlei and Rabbat, Michael and Ballas, Nicolas and LeCun, Yann and Bar, Amir and others},
  booktitle={Proceedings of the IEEE/CVF International Conference on Computer Vision},
  pages={370--382},
  year={2025}
}

@article{mur2026v,
  title={V-jepa 2.1: Unlocking dense features in video self-supervised learning},
  author={Mur-Labadia, Lorenzo and Muckley, Matthew and Bar, Amir and Assran, Mido and Sinha, Koustuv and Rabbat, Mike and LeCun, Yann and Ballas, Nicolas and Bardes, Adrien},
  journal={arXiv preprint arXiv:2603.14482},
  year={2026}
}

@article{heusel2017gans,
  title   = {Gans trained by a two time-scale update rule converge to a local nash equilibrium},
  author  = {Heusel, Martin and Ramsauer, Hubert and Unterthiner, Thomas and Nessler, Bernhard and Hochreiter, Sepp},
  journal = {Advances in neural information processing systems},
  volume  = {30},
  year    = {2017}
}

@inproceedings{zhang2018unreasonable,
  title   = {The unreasonable effectiveness of deep features as a perceptual metric},
  author  = {Zhang, Richard and Isola, Phillip and Efros, Alexei A and Shechtman, Eli and Wang, Oliver},
  booktitle = {Proceedings of the IEEE conference on computer vision and pattern recognition},
  pages   = {586--595},
  year    = {2018}
}

@article{unterthiner2018towards,
  title   = {Towards accurate generative models of video: A new metric \& challenges},
  author  = {Unterthiner, Thomas and Van Steenkiste, Sjoerd and Kurach, Karol and Marinier, Raphael and Michalski, Marcin and Gelly, Sylvain},
  journal = {arXiv preprint arXiv:1812.01717},
  year    = {2018}
}

@inproceedings{lin2024common,
  title={Common diffusion noise schedules and sample steps are flawed},
  author={Lin, Shanchuan and Liu, Bingchen and Li, Jiashi and Yang, Xiao},
  booktitle={Proceedings of the IEEE/CVF winter conference on applications of computer vision},
  pages={5404--5411},
  year={2024}
}

@misc{balestriero2025lejepaprovablescalableselfsupervised,
      title={LeJEPA: Provable and Scalable Self-Supervised Learning Without the Heuristics}, 
      author={Randall Balestriero and Yann LeCun},
      year={2025},
      eprint={2511.08544},
      archivePrefix={arXiv},
      primaryClass={cs.LG},
      url={https://arxiv.org/abs/2511.08544}, 
}

@misc{worldlabs2025marble,
  title        = {Marble: A Multimodal World Model},
  author       = {{World Labs}},
  year         = {2025},
  month        = nov,
  day          = {12},
  howpublished = {\url{https://www.worldlabs.ai/blog/marble-world-model}},
}

@article{shen2026lyra,
  title={Lyra 2.0: Explorable Generative 3D Worlds},
  author={Shen, Tianchang and Bahmani, Sherwin and He, Kai and Srinivasan, Sangeetha Grama and Cao, Tianshi and Ren, Jiawei and Li, Ruilong and Wang, Zian and Sharp, Nicholas and Gojcic, Zan and others},
  journal={arXiv preprint arXiv:2604.13036},
  year={2026}
}

@article{zhen2025tesseract,
  title={Tesseract: learning 4d embodied world models},
  author={Zhen, Haoyu and Sun, Qiao and Zhang, Hongxin and Li, Junyan and Zhou, Siyuan and Du, Yilun and Gan, Chuang},
  journal={arXiv preprint arXiv:2504.20995},
  year={2025}
}

@article{huang2026pointworld,
  title={PointWorld: Scaling 3D World Models for In-The-Wild Robotic Manipulation},
  author={Huang, Wenlong and Chao, Yu-Wei and Mousavian, Arsalan and Liu, Ming-Yu and Fox, Dieter and Mo, Kaichun and Fei-Fei, Li},
  journal={arXiv preprint arXiv:2601.03782},
  year={2026}
}

@article{savva2026solaris,
  title={Solaris: Building a multiplayer video world model in minecraft},
  author={Savva, Georgy and Michel, Oscar and Lu, Daohan and Waiwitlikhit, Suppakit and Meehan, Timothy and Mishra, Dhairya and Poddar, Srivats and Lu, Jack and Xie, Saining},
  journal={arXiv preprint arXiv:2602.22208},
  year={2026}
}

@article{hafner2025training,
  title={Training agents inside of scalable world models},
  author={Hafner, Danijar and Yan, Wilson and Lillicrap, Timothy},
  journal={arXiv preprint arXiv:2509.24527},
  year={2025}
}

@inproceedings{bar2025navigation,
  title={Navigation world models},
  author={Bar, Amir and Zhou, Gaoyue and Tran, Danny and Darrell, Trevor and LeCun, Yann},
  booktitle={Proceedings of the Computer Vision and Pattern Recognition Conference},
  pages={15791--15801},
  year={2025}
}

@article{chi2025wow,
  title={Wow: Towards a world omniscient world model through embodied interaction},
  author={Chi, Xiaowei and Jia, Peidong and Fan, Chun-Kai and Ju, Xiaozhu and Mi, Weishi and Zhang, Kevin and Qin, Zhiyuan and Tian, Wanxin and Ge, Kuangzhi and Li, Hao and others},
  journal={arXiv preprint arXiv:2509.22642},
  year={2025}
}

@inproceedings{kodaira2025streamdiffusion,
  title={Streamdiffusion: A pipeline-level solution for real-time interactive generation},
  author={Kodaira, Akio and Xu, Chenfeng and Hazama, Toshiki and Yoshimoto, Takanori and Ohno, Kohei and Mitsuhori, Shogo and Sugano, Soichi and Cho, Hanying and Liu, Zhijian and Tomizuka, Masayoshi and others},
  booktitle={Proceedings of the IEEE/CVF International Conference on Computer Vision},
  pages={12371--12380},
  year={2025}
}

@article{huang2026self,
  title={Self forcing: Bridging the train-test gap in autoregressive video diffusion},
  author={Huang, Xun and Li, Zhengqi and He, Guande and Zhou, Mingyuan and Shechtman, Eli},
  journal={Advances in Neural Information Processing Systems},
  volume={38},
  pages={167283--167308},
  year={2026}
}

@article{yang2025longlive,
  title={Longlive: Real-time interactive long video generation},
  author={Yang, Shuai and Huang, Wei and Chu, Ruihang and Xiao, Yicheng and Zhao, Yuyang and Wang, Xianbang and Li, Muyang and Xie, Enze and Chen, Yingcong and Lu, Yao and others},
  journal={arXiv preprint arXiv:2509.22622},
  year={2025}
}

@misc{maes_lelidec2026swm,
  title  = {stable-worldmodel-v1: Reproducible World Modeling Research and Evaluation},
  author = {Lucas Maes and Quentin Le Lidec and Dan Haramati and
            Nassim Massaudi and Damien Scieur and Yann LeCun and
            Randall Balestriero},
  year   = {2026},
  eprint = {2602.08968},
  archivePrefix = {arXiv},
  primaryClass = {cs.AI},
  url    = {https://arxiv.org/abs/2602.08968},
}

@article{
    mahajan2025jasmine,
    title={Jasmine: A simple, performant and scalable JAX-based world modeling codebase},
    author={Mihir Mahajan and Alfred Nguyen and Franz Srambical and Stefan Bauer},
    journal = {p(doom) blog},
    year={2025},
    url={https://pdoom.org/jasmine.html},
    note = {https://pdoom.org/blog.html}
}

@inproceedings{
    bruce2024genie,
    title={Genie: Generative Interactive Environments},
    author={Jake Bruce and Michael D Dennis and Ashley Edwards and Jack Parker-Holder and Yuge Shi and Edward Hughes and Matthew Lai and Aditi Mavalankar and Richie Steigerwald and Chris Apps and Yusuf Aytar and Sarah Maria Elisabeth Bechtle and Feryal Behbahani and Stephanie C.Y. Chan and Nicolas Heess and Lucy Gonzalez and Simon Osindero and Sherjil Ozair and Scott Reed and Jingwei Zhang and Konrad Zolna and Jeff Clune and Nando de Freitas and Satinder Singh and Tim Rockt{\"a}schel},
    booktitle={Forty-first International Conference on Machine Learning},
    year={2024},
    url={https://openreview.net/forum?id=bJbSbJskOS}
}

@misc{jax2018github,
  author = {James Bradbury and Roy Frostig and Peter Hawkins and Matthew James Johnson and Yash Katariya and Chris Leary and Dougal Maclaurin and George Necula and Adam Paszke and Jake Vander{P}las and Skye Wanderman-{M}ilne and Qiao Zhang},
  title = {{JAX}: composable transformations of {P}ython+{N}um{P}y programs},
  url = {http://github.com/jax-ml/jax},
  version = {0.3.13},
  year = {2018},
}
\newpage 

\newpage
\appendix

\end{document}